\documentclass[mathpazo,online]{jml}

\renewcommand\thisvolume{1}
\renewcommand\thisnumber{2}
\renewcommand\thisyear {2022}
\renewcommand\thismonth{June}

\renewcommand\originalreceiveddate{xx/xx/2022}

\renewcommand\acceptdate{xx/xx/2022}

\renewcommand\pagestart{1}
\renewcommand\pageend{xx}

\renewcommand\articledoi{doi.org/10.4208/jml.xxx}

\renewcommand\handler{xxx}

\newtheorem{observation}{Observation}[section]
\usepackage{lipsum}
\usepackage{xcolor}
\usepackage{colortbl}

\usepackage[utf8]{inputenc}
\usepackage{mathrsfs,amsthm,xcolor,verbatim,bbm,amsmath,amsfonts,
amssymb,nicefrac,enumitem}
\usepackage{hyperref,bm,mathtools,xparse,etoolbox}
\usepackage[capitalise,sort]{cleveref} 

\usepackage{textcomp}
\usepackage{sidecap}

\crefname{enumi}{item}{items}

\crefname{figure}{Figure}{Figures}
\crefname{equation}{}{}
\crefname{subsection}{Section}{Sections}

\crefname{case}{Case}{Cases}
\crefname{cor}{Corollary}{Corollaries}
\newcommand{\R}{\mathbb{R}}

\newcommand{\N}{\mathbb{N}}

\newcommand{\depen}[1]{^{(#1)}}

\newcommand{\rh}{\bm \rho\depen{\bm H}}

\newcommand{\X}{{\mathcal X}}
\newcommand{\Y}{{\mathcal Y}}

\newcommand{\alphamax}{\alpha_{\text{max}}}

\newcommand{\rhop}{\rho_{\text{poly}}}
\newcommand{\rhoe}{\rho_{\text{exp}}}
\newcommand{\rhoa}{\rho_{\text{Ai}}}
\newcommand{\rhoi}{\rho_{\delta}}

\newcommand{\muexp}{\mu_{\text{exp}}}
\newcommand{\mupoly}{\mu_{\text{poly}}}
\newcommand{\mudelta}{\mu_{\delta}}
\newcommand{\muAi}{\mu_{\text{Ai}}}

\DeclarePairedDelimiter{\norm}{\lVert}{\rVert}
\DeclarePairedDelimiter{\abs}{\lvert}{\rvert}

\usepackage{tikz}
\usetikzlibrary{matrix,chains,positioning,decorations.pathreplacing,arrows}
\usetikzlibrary{shapes,arrows}
\tikzset{
	font={\fontsize{9pt}{12}\selectfont}}

\ExplSyntaxOn

\bool_new:N \g_noteobserve

\NewDocumentCommand{\setnote}{}{
  \bool_gset_true:N \g_noteobserve
}

\NewDocumentCommand{\setobserve}{}{
  \bool_gset_false:N \g_noteobserve
}

\NewDocumentCommand{\nobs}{ o }{
  \IfValueT{#1}{
    \str_if_eq:noTF {note} {#1} {
      \bool_gset_true:N \g_noteobserve
    } {
      \str_if_eq:noTF {Note} {#1} {
        \bool_gset_true:N \g_noteobserve
      } {
        \bool_gset_false:N \g_noteobserve
      }
    }
  }
  \bool_if:nTF { \g_noteobserve } {
    \bool_gset_false:N \g_noteobserve
    note
  } {
    \bool_gset_true:N \g_noteobserve
    observe
  }
  \IfValueF{#1}{~}
}

\NewDocumentCommand{\Nobs}{ o }{
  \IfValueT{#1}{
    \str_if_eq:noTF {note} {#1} {
      \bool_gset_true:N \g_noteobserve
    } {
      \str_if_eq:noTF {Note} {#1} {
        \bool_gset_true:N \g_noteobserve
      } {
        \bool_gset_false:N \g_noteobserve
      }
    }
  }
  \bool_if:nTF { \g_noteobserve } {
    \bool_gset_false:N \g_noteobserve
    Note
  } {
    \bool_gset_true:N \g_noteobserve
    Observe
  }
  \IfValueF{#1}{~}
}

\int_new:N \g_furthermore

\NewDocumentCommand{\Moreover}{ o o }{
  \IfValueT{#1}{
    \str_case:nn {#1} {
      {Furthermore} {\int_set:Nn {\g_furthermore} {0}}
      {Moreover} {\int_set:Nn {\g_furthermore} {1}}
      {In~addition} {\int_set:Nn {\g_furthermore} {2}}
      {note} {\bool_gset_true:N \g_noteobserve}
      {observe} {\bool_gset_false:N \g_noteobserve}
    }
    \IfValueT{#2}{
      \str_case:nn {#2} {
        {Furthermore} {\int_set:Nn {\g_furthermore} {0}}
        {Moreover} {\int_set:Nn {\g_furthermore} {1}}
        {In~addition} {\int_set:Nn {\g_furthermore} {2}}
        {note} {\bool_gset_true:N \g_noteobserve}
        {observe} {\bool_gset_false:N \g_noteobserve}
      }
    }
  }
  \int_case:nn { \int_mod:nn {\g_furthermore} {3} } {
    { 0 } { Furthermore,~\nobs that}
    { 1 } { Moreover,~\nobs that}
    { 2 } { In~addition,~\nobs that}
  }
  \int_incr:N \g_furthermore
  \IfValueF{#1}{~}
}

\bool_new:N \g_hencetherefore

\NewDocumentCommand{\hence}{}{
  \bool_if:nTF { \g_hencetherefore } {
    \bool_gset_false:N \g_hencetherefore
    hence~
  } {
    \bool_gset_true:N \g_hencetherefore
    therefore~
  }
}

\NewDocumentCommand{\Hence}{}{
  \bool_if:nTF { \g_hencetherefore } {
    \bool_gset_false:N \g_hencetherefore
    Hence,~we~obtain~
  } {
    \bool_gset_true:N \g_hencetherefore
    Therefore,~we~obtain~
  }
}

\seq_new:N \g_cflist_loaded
\seq_new:N \g_cflist_pending

\NewDocumentCommand{\cfadd}{ m }
{
	\seq_if_in:NnF \g_cflist_loaded { #1 } {
		\seq_if_in:NnF \g_cflist_pending { #1 } {
			\seq_gput_right:Nn \g_cflist_pending { #1 }
		}
	}
}

\NewDocumentCommand{\cfconsiderloaded}{ m }{
	\seq_gput_right:Nn \g_cflist_loaded {#1}
}

\NewDocumentCommand{\cfremove}{ m }
{
	\seq_gremove_all:Nn \g_cflist_pending { #1 }
}

\NewDocumentCommand{\cfload}{ o }
{
	\seq_if_empty:NTF \g_cflist_pending {\unskip} {
		(cf.\ \cref{\seq_use:Nn \g_cflist_pending {,}})\IfValueTF{#1}{#1~}{\unskip}
		\seq_gconcat:NNN \g_cflist_loaded \g_cflist_loaded \g_cflist_pending
		\seq_gclear:N \g_cflist_pending
	}
}

\NewDocumentCommand{\cfclear} {} {
	\seq_gclear:N \g_cflist_loaded
	\seq_gclear:N \g_cflist_pending
}

\NewDocumentCommand{\cfout}{ o }
{
	\seq_if_empty:NTF \g_cflist_pending {\unskip} {
		(cf.\ \cref{\seq_use:Nn \g_cflist_pending {,}})\IfValueTF{#1}{#1~}{\unskip}
		\seq_gclear:N \g_cflist_pending
	}
}

\NewDocumentCommand{\ifnocf} { m } {
	\seq_if_empty:NT \g_cflist_pending { #1 }
}

\ExplSyntaxOff

\NewDocumentEnvironment{cproof}{m}
{\begin{proof}[Proof of \cref{#1}]}%
	{\noindent The proof of \cref{#1} is thus complete.
\end{proof}}

\NewDocumentEnvironment{cproof2}{m}
{\begin{proof}[Proof of \cref{#1}]}%
	{\noindent This completes the proof of \cref{#1}.
\end{proof}}

\newcommand{\revision}[1]{#1}

\begin{document}

\title{Numerical Investigation of Sequence Modeling Theory using Controllable Memory Functions}

\author[1,2]{
Haotian Jiang
	\thanks{
		{\tt haotian@nus.edu.sg}.
	}
}
\author[1]{
Zeyu Bao
	\thanks{
		{\tt zeyu@u.nus.edu}.
	}
}

\author[1]{
Shida Wang
	\thanks{
		{\tt e0622338@u.nus.edu}.
	}
}

\author[1,2]{
Qianxiao Li
	\thanks{
	Corresponding author.
{\tt qianxiao@nus.edu.sg}.
	}
}

\affil[1]{Department of Mathematics, National University of Singapore }
\affil[2]{Institute for Functional Intelligent Materials, National University of Singapore }

\begin{abstract}
The evolution of sequence modeling architectures, from recurrent neural networks and convolutional models to
  Transformers and structured state-space models, reflects ongoing efforts to address the diverse temporal dependencies
  inherent in sequential data. Despite this progress, systematically characterizing the strengths and limitations of these
  architectures remains a fundamental challenge. In this work, we propose a synthetic benchmarking framework to evaluate how
  effectively different sequence models capture distinct temporal structures. The core of this approach is to generate
  synthetic targets, each characterized by a parametric memory function $\rho(s, \alpha)$ and a controllable parameter $\alpha$ that determines
  the temporal strength. This setup allows us to produce a continuum of tasks that vary in temporal complexity,
  enabling fine-grained analysis of model behavior with respect to specific memory properties. We focus on four representative
  memory functions, each corresponding to a distinct class of temporal structures: exponential and polynomial functions for
  decay dynamics, impulse functions for long-range dependencies, and Airy functions for sparsity patterns. Experiments on
  several sequence modeling architectures confirm existing theoretical insights and reveal new findings regarding approximation
   capabilities, optimization dynamics, and architectural trade-offs. These results demonstrate the effectiveness of the
  proposed method in advancing theoretical understanding and highlight the importance of using controllable targets with
  clearly defined structures for evaluating sequence modeling architectures.
\end{abstract}

%
%
%
\keywordone{xxx,}
\keywordtwo{xxx, }
\keywordthree{xxx, }
\keywordfour{xxx.}
\keywordfive{}

\maketitle

\section{Introduction}\label{sec:introduction}

Sequence modeling is a fundamental task in machine learning,
with applications spanning natural language processing (NLP),
time-series forecasting,
weather prediction, and the modeling of dynamical systems \cite{song2024.DeepLearningbasedTime, kong2025.DeepLearningTime, pinheiro2025.InterpretableMachineLearning,chattopadhyay2020.DatadrivenPredictionsMultiscale,gajamannage2023.RecurrentNeuralNetworks}.
In particular, large language models (LLMs) such as GPT \cite{radford2018.ImprovingLanguageUnderstanding,brown2020.LanguageModelsAre}, Claude \cite{anthropic2024.Claude3Model}, and Gemini \cite{geminiteam2023.GeminiFamilyHighly} have demonstrated remarkable capabilities across a wide range of language understanding and generation tasks, fundamentally advancing the field of NLP and highlighting the critical importance of sequence modeling architectures.
These tasks often involve capturing diverse temporal structures,
which require specialized architectures to model the underlying patterns effectively.
Over the years, a variety of sequential architectures have been proposed, each with different strategies for handling temporal dependencies.

Recurrent neural networks (RNNs) \cite{rumelhart1986.LearningRepresentationsBackpropagating} were among the earliest architectures designed for learning sequential patterns.
Despite their foundational role, RNNs are known to struggle with long-range dependencies due to issues such as vanishing gradients \cite{bengio1994.LearningLongtermDependencies,kolen2001.GradientFlowRecurrent,pascanu2013.DifficultyTrainingRecurrent}. 
To mitigate these limitations, gated recurrent architectures were introduced, including Long Short-Term Memory (LSTM) networks \cite{hochreiter1997.LongShortTermMemory} and Gated Recurrent Units (GRUs) \cite{cho2014.LearningPhraseRepresentations}, which incorporate mechanisms to better manage information over extended time horizons.
Beyond recurrent designs, convolutional architectures such as WaveNet \cite{oord2016.WaveNetGenerativeModel} and Temporal Convolutional Networks (TCNs) have demonstrated strong performance on sequence tasks.
By using dilated causal convolutions, these models can capture long-range temporal dependencies efficiently.
Empirical studies \cite{bai2018.EmpiricalEvaluationGeneric} have shown that convolutional models can outperform recurrent ones in various settings.
The Transformer architecture \cite{vaswani2017.AttentionAllYou} introduced a fundamentally different approach by employing self-attention, allowing direct interaction between all elements of a sequence regardless of distance.
This has enabled major advances in NLP, making Transformers the dominant architecture in the field.
However, the computational cost of self-attention grows quadratically with sequence length, limiting scalability for extremely long sequences.
More recently, state-space models (SSMs) have emerged as a promising direction, drawing on principles from dynamical systems theory.
Models such as the Structured State Space model (S4) \cite{gu2022.EfficientlyModelingLong}, S4D \cite{gu2022.ParameterizationInitializationDiagonal}, and Linear Recurrent Units (LRU) \cite{orvieto2023.ResurrectingRecurrentNeural} are proposed to handle long sequences efficiently.

Despite the variety of available architectures, systematically understanding their strengths and limitations remains challenging.
Most current evaluations rely on large, real-world benchmarks, such as Long Range Arena (LRA) \cite{tay2020.LongRangeArena} and LongBench \cite{bai2024.LongBenchBilingualMultitask}.
Although these benchmarks are useful for measuring overall performance, they suffer from a fundamental limitation: the temporal structures in these datasets are complex and undefined, making it difficult to identify which temporal properties contribute to performance differences.
While several theoretical studies have explored how different architectures handle temporal dependencies, such as analyzing linear RNNs (equivalently shallow state-space models) through impulse response \cite{li2022.ApproximationOptimizationTheory} or examining approximation capabilities of nonlinear RNNs, convolutional architectures, and Transformers \cite{li2021.ApproximationPropertiesRecurrent, wang2023.InverseApproximationTheory, jiang2021.ApproximationTheoryConvolutional, jiang2024.ApproximationRateTransformer}, these works highlight that each architecture is intrinsically suited to capturing distinct temporal structures.
Nonetheless, a systematic approach for directly testing these theoretical properties across different models through controlled experiments is lacking. What is needed is a principled methodology that isolates individual temporal characteristics, enabling us to probe each architecture's intrinsic capabilities in a controlled setting.

To bridge this gap, we propose a systematic evaluation methodology using synthetic sequence targets with mathematically well-defined temporal dependencies.
The core of this approach is the use of parametric memory functions $\rho(s, \alpha)$, which provide a natural mathematical characterization of how past inputs influence current outputs.
As established in prior theoretical work \cite{li2022.ApproximationOptimizationTheory, jiang2021.ApproximationTheoryConvolutional}, memory functions uniquely determine temporal dependency structures, making them an ideal foundation for systematic evaluation.
Each target is generated from such a memory function, where the parameter $\alpha$ controls the temporal strength.
By varying $\alpha$, we create a continuum of tasks with varying temporal complexity, enabling fine-grained analysis of model behavior.
This framework allows us to interpret model performance more precisely:

\begin{enumerate}
    \item if the error increases rapidly with $\alpha$, the model struggles with stronger temporal dependencies;
    \item if the error remains stable, the model handles the structure efficiently;
    \item if the error fluctuates across different seeds, the model is sensitive to initialization or stochastic factors such as optimization dynamics or weight initialization.

\end{enumerate}

We focus on four representative memory functions that systematically capture different aspects of temporal structure: exponential and polynomial decay (characterizing decay dynamics), impulse (characterizing long-range dependencies), and Airy (characterizing sparsity patterns).
Compared to existing benchmark methods, our approach provides clear, controllable temporal structures and is versatile enough to evaluate both small and large architectures.
It also offers a foundation for further theoretical analysis, including approximation, optimization, and generalization properties, making it a flexible and principled tool for understanding sequential architectures.

Experiments on LSTM, S4D, TCN, and Transformer architectures both validate established theoretical predictions and uncover new phenomena.
For recurrent and state-space models, we confirm their sensitivity to memory decay rates, while revealing that depth provides benefits for sparse dependencies but degrades performance on smoothly-decaying structures.
For convolutional architectures, we demonstrate stable performance on sparse long-range dependencies and identify a novel tail-energy complexity measure that better characterizes approximation difficulty than binary sparsity.
For Transformers, we validate effective rank theory explanations and reveal critical trade-offs between attention head count and per-head dimensionality that depend on the temporal structure.
Furthermore, we demonstrate that sequential composition of architectures can substantially compensate for individual weaknesses, with mixed models achieving performance close to the stronger component architecture.
These results highlight the effectiveness of controllable synthetic targets for understanding sequence modeling capabilities and provide practical insights for architecture selection and design.



\section{Formulation of the targets}\label{sec: memory function definition}

In this section,
we establish the mathematical framework for characterizing temporal relationships in sequence modeling through memory functions.
We proceed in four steps.
First (\cref{sec: preliminaries}), we formulate sequential targets as mappings between function spaces and prove that linear, causal, time-homogeneous targets admit unique memory function representations.
Second (\cref{sec: parametric memory}), we introduce the parametric memory function framework that enables controllable evaluation.
Third (\cref{sec: four memory functions}), we define four representative memory functions designed to probe different temporal structures.
Finally (\cref{sec: synthetic construction}), we specify the synthetic target construction and justify the extension to nonlinear targets.

\subsection{Memory Function Representation for Linear Sequential Targets} \label{sec: preliminaries}
We discuss the theoretical preliminaries underlying the definition of sequential targets,
building upon the theoretical framework from \cite{li2022.ApproximationOptimizationTheory,jiang2021.ApproximationTheoryConvolutional}.
Sequential targets are formulated as mappings between function spaces,
enabling a precise mathematical representation of the input-output relationships in sequence modeling.

\paragraph{Notations} Throughout this paper, we use boldface symbols to denote sequences or operators (e.g., $\bm{x}$, $\bm{y}$, $\bm{H}$, $\bm{\rho}$), while non-boldface symbols with arguments denote the value at specific indices (e.g., $x(t)$, $y(t)$, $\rho(s)$). We use non-boldface with subscript $t$ to denote the functional at time $t$ (e.g., $H_t$).

For concreteness, we consider discrete sequences and scalar input-output.
Specifically,
we define the input space $\X$ and the output space $\Y$ as: 
\begin{align}
  \text{Input space: }  \X &= \{ \bm x: x(t)\in \R \text{ for } t\in \N_{\geq 0}\}, \\
   \text{Output space: } \Y &= \{ \bm y: y(t)\in \R \text{ for } t\in \N_{\geq 0}\}.
\end{align}
 Both input sequence $\bm{x}$ and output sequence $\bm{y}$ can be regarded as functions defined over the index set $\N_{\geq 0}$.
A sequential target is defined as a mapping $\bm{H}: \X \to \Y$, where $\bm{H}$ can be considered as an operator between $\X$ and $\Y$.
For each time point $t$, the input-output relationship can be expressed as:
\begin{align}
    y(t) = H_t(\bm{x}),
\end{align}
where $ H_t: \X \to \mathbb{R} $ is treated as a functional, mapping the entire input sequence $\bm{x}$  to a scalar output $y(t)$. Importantly, the output $ y(t) $ may depend on the entire input sequence $ \bm{x} $, capturing the temporal dependencies inherent in the target.

To characterize temporal dependencies in sequential targets, we introduce the concept of a memory function, which describes how outputs are determined by past inputs.
We show that a memory function can be uniquely defined for linear targets satisfying four natural properties: linearity, continuity, causality, and time-homogeneity.
We first define these properties for the sequential target $\bm H$.

\begin{itemize}
    \item \textbf{Linearity \& Continuity: }
    $H_t$ is a continuous linear functional if for any $\bm x_1,  \bm x_2 \in \mathcal X$
	and $\lambda_1, \lambda_2 \in \mathbb R$,
	\begin{equation}
		\begin{gathered}
		H_t(\lambda_1\bm x_1 + \lambda_2\bm x_2) = \lambda_1 H_t(\bm x_1) + \lambda_2 H_t(\bm x_2), \\
		\norm{H_t} := \sup_{\bm x \in \mathcal X, \norm{\bm x}_{\mathcal X}\leq 1}\abs{H_t(\bm x)} < \infty,
		\end{gathered}
	\end{equation}
	where $\norm{\bm x}_{\mathcal X}$ is the norm of the input space and $\norm{H_t}$ denotes the induced functional norm. 

    \item \textbf{Causality: }
    $\bm H$ is \emph{causal} if it only depends on past inputs:
	for any $\bm x_1,  \bm x_2 \in \mathcal X$ and any $t \in \N_{\geq 0}$ such that
	$
		x_1(s) = x_2(s) \ \text{ for all } s \leq t
	$,
	the output satisfies $H_t(\bm x_1) = H_t(\bm x_2)$.

    \item \textbf{Time-homogeneity:}
    For a sequence $\bm x \in \X$, a shift of $\bm x$ by $\tau$ is denoted as 
    \begin{equation}
        x\depen{\tau}(s) :=
        \begin{cases}
             x(s-\tau)  &  s \geq \tau,\\
             0           & 0\leq s< \tau.
        \end{cases}
    \end{equation}
    
    $\bm H$ is \emph{time-homogeneous} if 
        for any $t, \tau \in \N_{\geq 0}$, we have
        $
            H_t(\bm x) = H_{t+\tau}(\bm x\depen{\tau}).
        $
        
    This is a similar idea to a time-invariant system in the context of dynamical systems,
    where the target relationship does not explicitly depend on time (see, e.g., \cite{oppenheim1996.SignalsSystems}).
\end{itemize}

Causality and time-homogeneity are fundamental temporal structures inherent to many sequential architectures. 
Examples include RNNs, GRUs, SSMs, TCNs, and Transformers equipped with relative position encoding and causal masking, 
all of which adhere to these principles.

The following result shows that associated with each sequential target operator $\bm H$
is a unique memory function, and its proof is found in Appendix \ref{appen: proof}.

\begin{theorem}\label{thm: linear target representation}
Let $\bm x\in\X$ with $\displaystyle\sum\abs{x}^2<\infty$ and $\norm{\bm x}_{\mathcal X} = \sqrt{\sum\abs{x}^2}$.
Define $\bm y \in \Y$ such that $y(t)=H_t(\bm x)$, where $\bm H$ is linear, continuous, causal, and time-homogeneous.
Then, there exists a unique function $\rh :\mathbb N_{\geq 0} \to \mathbb R$ such that for all $x\in\X$ we have
    \begin{equation}\label{eq: linear target representation}
    	H_t(\bm x) = \sum_{s=0}^{t} \rho\depen{\bm H}(s) x(t-s), \quad t \in \mathbb \N_{\geq 0}.
    \end{equation}
    This $\bm \rho\depen{\bm H}$ is called the \emph{memory function} of $\bm H$,
    and satisfies $\displaystyle\sum_t \abs{\bm\rho\depen{\bm H}}(t)^2<\infty$.
\end{theorem}

The specific structure of $\bm\rho\depen{\bm H}$ determines targets with distinct temporal properties.
For instance, \cite{li2022.ApproximationOptimizationTheory} shows that RNN performance is closely tied to the decay rate and smoothness of $\bm\rho\depen{\bm H}$, while \cite{jiang2021.ApproximationTheoryConvolutional} demonstrates that TCN performance is influenced by the sparsity of $\bm\rho\depen{\bm H}$.
These findings highlight how the memory function plays a central role in determining which architectures are effective at capturing different temporal structures.

\revision{
\begin{remark}[Continuous-time extension and irregular sampling]
Theorem~\ref{thm: linear target representation} is stated for discrete uniformly sampled sequences for concreteness and controlled evaluation. 
The memory function representation extends naturally to continuous time, where the target takes the integral form
\begin{equation}
    H_t(x) = \int_0^t \rho(s)\, x(t-s)\, ds,
\end{equation}
and an analogous representation theorem holds under the same conditions of linearity, continuity, causality, and time-homogeneity. Under this continuous-time formulation, the memory function $\rho(s)$ is a characterization of the underlying system and is independent of the sampling grid. Irregular sampling therefore does not alter the memory structure itself, it only affects how architectures access it at observed time points. The discrete uniform benchmark studied in this paper is a deliberate controlled instantiation of the continuous-time framework, chosen to isolate temporal structure in a precise and interpretable way.
\end{remark}

}

\subsection{Parametric Memory Functions}\label{sec: parametric memory}

Motivated by the theoretical insights above,
we leverage the memory function representation to design a systematic benchmarking framework.
\cref{sec: preliminaries} has shown that a linear target can be uniquely determined when given a memory function $\bm\rho\depen{\bm H}$.
By constructing functions $\rho(s, \alpha)$ with a controllable parameter $\alpha$, we can generate a continuum of synthetic targets that isolate specific temporal properties.
For instance, by varying $\alpha$, we can control decay rate, dependency range, or sparsity, enabling systematic evaluation of how different architectures respond to varying temporal strengths.

We refer to $ \rho(s,\alpha) $ as a \textbf{parametric memory function}, which maps a discrete time index $ s $ and a controllable parameter $ \alpha $ to a real-valued weight:
\begin{equation}
    \rho: \mathbb{N}_{\geq 0} \times [0, 1] \rightarrow \mathbb{R}.
\end{equation}
The parameter $ \alpha \in [0,1]$ controls the temporal strength,
with larger values corresponding to stronger memory effects.
For practical implementation considerations, 
we use strictly increasing scaling functions $\mu(\alpha)$ to map $\alpha$ to appropriate ranges for each specific memory function; these design choices are discussed in \cref{sec: scaling functions}.

Having established the general parametric framework, we now define four representative memory functions, each designed to probe a distinct aspect of temporal structure.

\subsection{Four Representative Memory Functions}\label{sec: four memory functions}

We consider four representative memory functions with distinct temporal characteristics:
\textbf{exponential} and \textbf{polynomial} decay ($\rhoe$ and $\rhop$, capturing memory decay speed),
\textbf{impulse} ($\rhoi$, capturing finite dependency range),
and \textbf{Airy} ($\rhoa$, capturing oscillatory behavior and sparsity).
Their key properties are summarized in \cref{table: memory functions}.

\begin{table}[h]
\centering
\caption{Types of memory functions $\rho$}
\begin{tabular}{|c||c|c|c|}
\hline
Name  & Smoothness & Decay & Sparsity \\
\hline
$\rhoe$   & smooth     &  controlled by $\alpha$        & high\\
$\rhop$    & smooth     &  controlled by $\alpha$     & high\\
$\rhoi$      & non-smooth &  decay         & low \\
$\rhoa$         & smooth     &  decay      & controlled by $\alpha$ \\
\hline
\end{tabular}\label{table: memory functions}
\end{table}

\paragraph{Exponential and Polynomial Decay}
The exponential and polynomial memory functions model decaying temporal dependencies, where recent inputs have greater influence than earlier ones, differing primarily in their decay rates.  

The exponential memory function is defined as:
\begin{equation}
    \rhoe(s, \alpha) = \exp\left(-\frac{s}{\muexp(\alpha)}\right), \quad \muexp(\alpha):[0,1]\to \R_{>0}.
\end{equation}

The polynomial memory function is defined as:
\begin{equation}
    \rhop(s, \alpha) = (1 + s)^{-\frac{1}{\mupoly(\alpha)}}, \quad \mupoly(\alpha):[0,1]\to \R_{>0}.
\end{equation}

As $\alpha$ increases, the decay becomes slower,  
resulting in stronger dependencies on past inputs.

\paragraph{Impulse}

In contrast to decay-based functions,
the impulse memory function captures long-range dependencies,
where the output depends on a specific distant input.
The impulse memory function is defined as:
\begin{align}
    \rhoi(s, \alpha) = 
    \begin{cases}
        1 & \text{if } s = \mudelta(\alpha), \\
        0 & \text{otherwise}
    \end{cases}, \quad \mudelta(\alpha):[0,1]\to \N_{\geq 0}.
\end{align}

In this case, the output at time $t$ depends solely on a single input value at a specific offset: 
$
    y(t) = x\big(t - \mudelta(\alpha) \big).
$
As $\alpha$ increases, the output is determined by inputs further in the past,  
capturing the effect of long-range memory.

\paragraph{Airy}

In addition to decay speed and dependency range,
we consider sparsity as a critical factor in temporal structure.
Following \cite{jiang2021.ApproximationTheoryConvolutional}, sparsity for memory functions is typically defined using a binary criterion based on the number of nonzero values.
To enable controllable sparsity via $\alpha$, we use a truncated version of the Airy function of the first kind, whose effective sparsity can be varied through shifting.
We define the truncated Airy function as
\begin{align}
    \text{Ai}_\text{tr}(s) = \begin{cases}
        \frac{1}{\pi} \int_0^\infty \cos\left( \frac{v^3}{3} + vs \right) \, dv & s < 20 \\
        0 & s \geq 20
    \end{cases} \, ,
\end{align}
where the truncation at $s = 20$ exploits the exponentially fast decay of the Airy function for positive $s$.  
Using this, we define the memory function as
\begin{equation}
    \rhoa(s,\alpha) = \text{Ai}_\text{tr}\big(\muAi\left(s - c \cdot \alpha\right)\big),\quad \muAi: [0,1]\to \R_{\geq 0},
\end{equation}
where $c > 0$ is a scaling factor.  
As $\alpha$ increases, the nonzero region of the function shifts outward,  
reducing the effective sparsity of the memory structure.

\begin{figure}[!ht]
    \centering
    \includegraphics[width=0.85\linewidth]{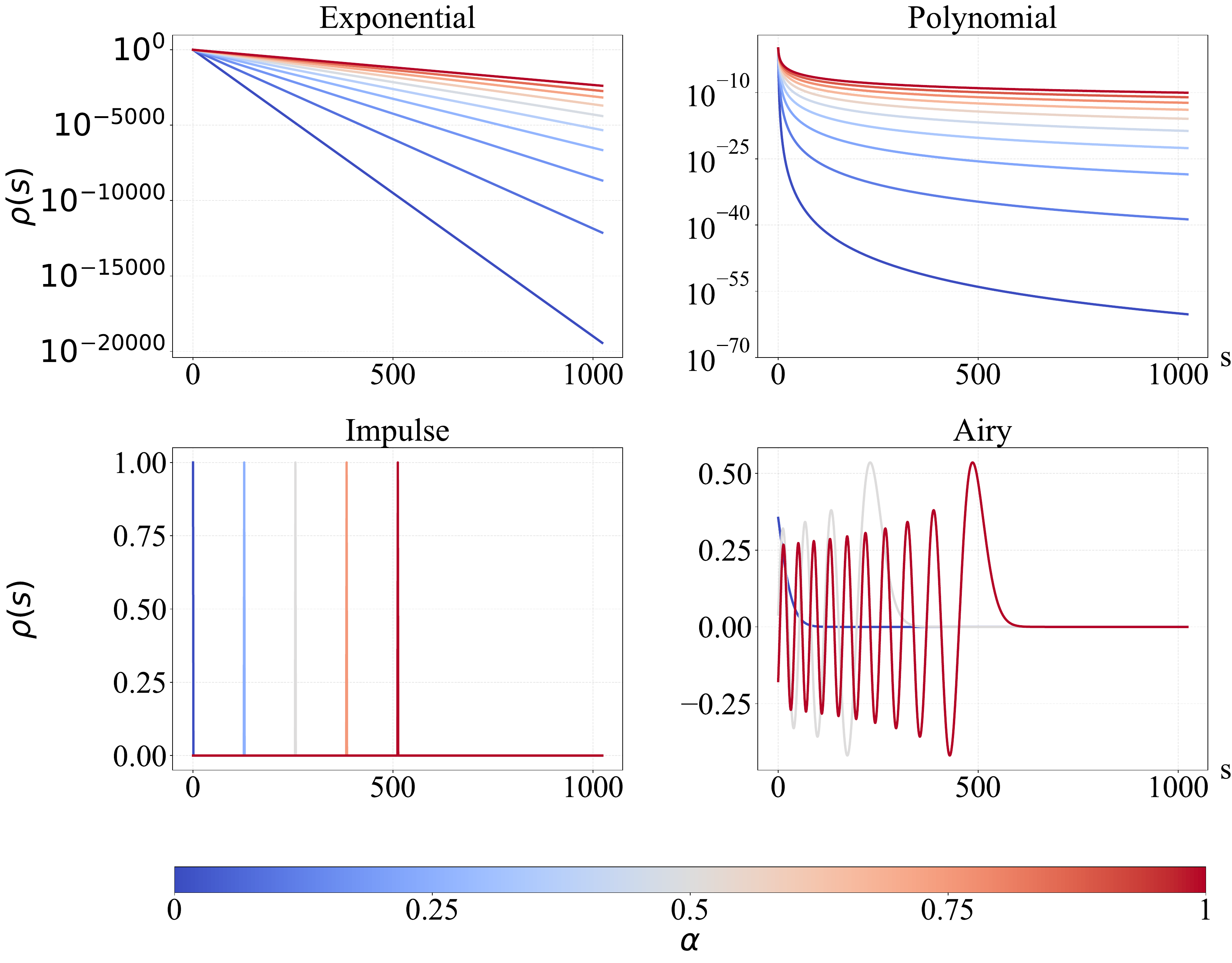}
    \caption{Plot of individual memory functions for $T=512$, illustrating variations with respect to the parameter $\alpha$.
The colorbar indicates the corresponding $\alpha$ values.
The scaling functions used for these plots are defined in \cref{sec: scaling functions}.
    }
    \label{fig: memory function plot}
\end{figure}

\cref{fig: memory function plot} visualizes these four memory functions across varying $\alpha$ values, illustrating how each function's temporal structure evolves with temporal strength.

\subsection{Synthetic Target Construction and Nonlinear Extension}\label{sec: synthetic construction}

Having defined the parametric memory functions for linear targets, we now extend the framework to nonlinear targets for evaluating sequence models.
While Theorem~\ref{thm: linear target representation} establishes that linear targets admit a unique memory function representation,
practical sequence modeling tasks often involve nonlinearity.
We therefore introduce activation functions $\sigma_1$ and $\sigma_2$ into the construction, using the parametric function $\rho(s, \alpha)$ as a temporal weighting function.
The synthetic target is generated as:
\begin{align}\label{eq: data generation}
    y(t) = H_t^{(\alpha)}(\bm{x}) := \sigma_1\left(\sum_{s=0}^{t} \rho(s, \alpha) \, \sigma_2\big(x(t - s)\big)\right), \quad t \in \{0, \dots, T\},
\end{align}
where $ \bm{x} $ is the input sequence, $ T $ is the maximum sequence length, $ \sigma_1, \sigma_2 $ are activation functions, and $ s $ represents the time lag.

\paragraph{Theoretical justification for nonlinear extension}\label{subsec: nonlinear justification}

We now show that $\rho(s, \alpha)$ continues to control the temporal dependency structure in the nonlinear construction \cref{eq: data generation}.
To do this, we first establish a general definition of memory function that applies to both linear and nonlinear targets.
For any target $\bm{H}$ that is causal and time-homogeneous,
we define the memory function by measuring the influence of past inputs on current outputs through the partial derivative
\begin{equation}\label{eq: memory derivative}
    \frac{\partial H_t(\bm{x})}{\partial x(s)}.
\end{equation}
This derivative-based definition is general and applies regardless of linearity.
For linear targets satisfying the conditions of \cref{thm: linear target representation},
this definition coincides with the memory function $\rho^{(\bm H)}$ from the representation theorem.
To see this,
consider $\displaystyle H_t(\bm{x}) = \sum_{k=0}^{t} \rho^{(\bm{H})}(k) x(t-k)$.
For any input time $s \in \{0, \ldots, t\}$,
the derivative with respect to $x(s)$ is
\begin{equation}
    \frac{\partial H_t(\bm{x})}{\partial x(s)} = \rho^{(\bm{H})}(t-s),
\end{equation}
since only the term $\rho^{(\bm{H})}(t-s) x(s)$ in the sum contains $x(s)$.
This shows that $\rho^{(\bm{H})}(\ell)$ measures the influence of an input that is $\ell$ time steps in the past.

Time-homogeneity allows us to use a canonical measurement: the influence of an input $s$ steps in the past is the same regardless of absolute time.
Thus, measuring the influence of $x(0)$ on $H_s(\bm{x})$ consistently captures the memory at lag $s$.
Since $\displaystyle H_s(\bm{x}) = \sum_{k=0}^{s} \rho^{(\bm{H})}(k) x(s-k)$, we have
\begin{equation}
    \frac{\partial H_s(\bm{x})}{\partial x(0)} = \rho^{(\bm{H})}(s).
\end{equation}
This justifies using $\rho^{(\bm{H})}(s) = \frac{\partial H_s(\bm{x})}{\partial x(0)}$ as the definition of memory function at lag $s$ for general targets.

We now apply this definition to the nonlinear targets in \cref{eq: data generation}:
\begin{equation}
    y(t) = \sigma_1\left(\sum_{s=0}^{t} \rho(s, \alpha) \, \sigma_2\big(x(t - s)\big)\right).
\end{equation}
Computing the derivative with respect to $x(0)$ gives us the memory function for this nonlinear target:
\begin{equation}\label{eq: nonlinear memory derivative}
    \frac{\partial y(t)}{\partial x(0)} = \sigma_1'\left(\sum_{k=0}^{t} \rho(k,\alpha)\sigma_2\big(x(t-k)\big)\right) \cdot \rho(t, \alpha) \cdot \sigma_2'\big(x(0)\big),
\end{equation}
where $\sigma_1'$ and $\sigma_2'$ denote the derivatives of the activation functions.
Crucially, the memory function (left-hand side) has $\rho(t, \alpha)$ as a multiplicative factor, showing that the temporal weighting function $\rho(t, \alpha)$ directly controls the memory structure.

When the activation functions are Lipschitz continuous with bounded derivatives,
the memory structure is preserved in a particularly clean form.
Specifically,
if there exist constants $0 < c_1 \leq \sigma_1'(z) \leq C_1$ and $0 < c_2 \leq \sigma_2'(z) \leq C_2$ for all $z$ in the relevant domain,
then the influence in the nonlinear case satisfies
\begin{equation}
    c_1 c_2 \, \abs{\rho(t, \alpha)} \leq \abs*{\frac{\partial y(t)}{\partial x(0)}} \leq C_1 C_2 \, \abs{\rho(t, \alpha)}.
\end{equation}
This shows that the memory function for the nonlinear target is dominated by $\rho(t, \alpha)$ up to a multiplicative scaling factor controlled by the activation function derivatives.
Importantly, the temporal structure and relative decay pattern encoded in $\rho(s, \alpha)$ are preserved.

This analysis shows that for the nonlinear targets constructed via \cref{eq: data generation}, the memory function (defined by the derivative \cref{eq: memory derivative}) is dominated by the temporal weighting function $\rho(s, \alpha)$.
Therefore, our benchmarking framework can systematically evaluate how architectures handle different temporal structures by varying $\rho(s, \alpha)$, with the nonlinear activations adding realistic complexity while the temporal dependencies remain controlled by $\rho(s, \alpha)$.

\section{Experimental Methodology and Setup}\label{sec:methodology}

In this section, we describe the methodology for evaluating how different temporal structures affect the theoretical properties of sequential architectures.
We first introduce our evaluation framework and the scaling methodology used to ensure fair comparison across memory functions.
Then we specify the complete experimental settings, including data generation, model parameters, and training procedures.

\subsection{Evaluation Framework}

Approximation, optimization, and generalization are three key aspects in machine learning theory.
\textbf{Approximation} refers to a model's ability to accurately capture a target's underlying structure.
In practice, we use the final training loss to estimate the approximation error bound,
under the condition that the model does not overfit.
\textbf{Optimization} focuses on the model's ability to effectively minimize the loss function during training.
This can be evaluated by measuring the number of training steps required to reach a given loss threshold,
and by examining the distribution of final losses across different random seeds to assess sensitivity to initialization.
\textbf{Generalization}, which refers to the model's ability to perform well on unseen data,
can be studied by varying the size of the training dataset and measuring how the test loss changes as a function of data availability.

In this work, we primarily focus on the approximation aspect due to the availability of existing theoretical studies, but we also present some interesting phenomena observed in optimization.
A detailed analysis of generalization is deferred to future work.

The core idea of our approach is to analyze how the performance of a model is influenced by different target temporal structures.
To do this, we fix the model size and train on targets generated with different values of $\alpha$ for each memory function.
This allows us to systematically assess how the model's properties,
such as its approximation error and its sensitivity to initialization,
change as the temporal strength increases.

For example, to evaluate approximation properties,
we can plot the loss as a function of $\alpha$,
after first ensuring that no overfitting occurs,
so that the loss accurately reflects the model's approximation error.
Comparing these curves across different memory functions reveals cases where a model is more influenced by one type of temporal structure than another.
This approach also allows for qualitative comparisons between models,
revealing which architectures are better suited for particular temporal structures and which struggle under certain conditions.

Overall, this framework enables both quantitative and qualitative comparisons of sequence models' theoretical properties,
helping validate existing theoretical insights while also revealing new empirical phenomena that may guide future theoretical developments.

\paragraph{Numerical Estimation of Approximation Error}
In supervised learning, the approximation error for a target $y$ within a hypothesis space $\mathcal{H}$ is defined as
\begin{align}
    e = \inf_{\hat y \in \mathcal{H}} \norm{y - \hat y},
\end{align}
which represents the minimum achievable distance between the target function and the hypothesis space.

\revision{
To numerically estimate this approximation error bound,
we use the final training loss as an upper bound.
This is valid under the condition that no overfitting occurs.
Since our targets are synthetically generated, we can produce arbitrarily large amounts of training data, which effectively prevents overfitting.
We verify this by checking that the test loss remains consistent with the training loss.
For the parameter settings used in this study, we have confirmed that no significant overfitting is observed (see \cref{appendix:val_curves}).

Additionally, to account for the stochastic nature of the training process,
we train each model on the same target using multiple random seeds.
We then take the minimum training loss across these runs as an estimate of the approximation error bound,
providing a more robust measure of model performance.

}

\paragraph{Quantifying Sensitivity to Temporal Strength}\label{para:alpha_sensitivity}
To provide a quantitative summary of how sensitive each architecture is to changes in temporal strength,
we compute the relative loss increment from the minimum to maximum $\alpha$ value for each memory function.
Specifically, for each model size $m$, we first compute the minimum training loss $L_{\alpha}$ across all random seeds for a given $\alpha$.
We then calculate the relative loss increment as
\begin{equation}
\max\left\{\frac{L_{\alpha_{\max}} - L_{\alpha_{\min}}}{L_{\alpha_{\min}}}, 0\right\},
\end{equation}
where the max operation ensures the metric is non-negative.
If the loss decreases with larger $\alpha$ (yielding a negative value), we set the metric to zero, indicating that the model is not affected by the increased temporal strength, since the increased task difficulty does not result in a loss increment.

Since the loss values can span multiple orders of magnitude,
we report the geometric mean of this ratio across different model sizes,
along with the standard error of the mean.
This metric quantifies how much the approximation difficulty increases as the temporal dependency strength grows,
enabling direct numerical comparisons across different architectures and memory functions.

\subsection{Scaling Methodology for Proper Comparison}\label{subsec: scaling}

In this subsection, we discuss several practical considerations for implementing the proposed methodology,
including scaling function definitions and loss scaling.
These are critical to ensuring consistent and meaningful evaluation across different memory functions,
sequence lengths, and the temporal strength parameter $\alpha$.

\paragraph{Scaling Function Definitions}\label{sec: scaling functions}

Recall from \cref{sec: memory function definition} that each type of memory function is paired with a scaling function $\mu$
that maps the parameter $\alpha\in[0,1]$ to an appropriate range, controlling the temporal strength.
These scaling functions are not arbitrary and must be carefully chosen to ensure that the target retains the intended temporal structure,
even for different sequence lengths $T$ used in practice.

Since both the exponential memory function $\rhoe$ and the polynomial memory function $\rhop$ exhibit decay over time,
it is important to compare them on a consistent scale.
To achieve this, we define the scaling function $\muexp(\alpha)$ such that for all $\alpha$, both functions have the same total area over $[0, T]$,
ensuring that their relative decay rates can be directly compared:
\begin{align}
    \sum_{s=0}^{T} \rhoe(s, \muexp(\alpha))  = \sum_{s=0}^{T} \rhop(s, \mupoly(\alpha)) .
\end{align}
This normalization removes the effect of the overall scale,
allowing us to isolate and compare the impact of different decay speeds accurately.

For the polynomial memory function $\rhop$, the scaling function is defined as
\begin{equation}
    \mupoly(\alpha) = \alpha \cdot \alphamax,
\end{equation}
where $\alphamax$ is chosen such that the tail contribution is negligible:
\begin{equation}
    \displaystyle \sum_{s=T+1}^{\infty} \rhop(s, \alphamax) = \epsilon,
\end{equation}
with $\epsilon$ being a small threshold (e.g., $10^{-8}$).

For the impulse memory function $\rhoi$,
we define the scaling function as
$\displaystyle \mudelta(\alpha) = \left\lfloor \frac{\alpha T}{2} \right\rfloor$,
which ensures that the maximum shifted distance $\mudelta(1)$ does not exceed half of the input sequence length.
Since the output is truncated at $T$, limiting the shift to $T/2$ ensures that at least half of the output sequence contains meaningful information (rather than zero-padding), allowing the model to effectively learn the target dependencies.

For the Airy memory function, we aim to control sparsity by shifting the oscillatory pattern.
We define the scaling as
$\displaystyle \rho_{\text{Ai}}(s,\alpha) = \text{Ai}_{\text{tr}}\left(40\left(\frac{s}{T} - \frac{\alpha}{2}\right)\right)$,
where the factor of 40 normalizes the function's period relative to the sequence length, and the shift by $\alpha/2$ ensures that as $\alpha$ increases from 0 to 1, the high-magnitude region shifts from the beginning to the middle of the sequence, reducing the effective sparsity.

\paragraph{Loss Scaling}

When comparing models on the same memory function across different $\alpha$ values, the output scale can vary significantly.
For example, the impulse memory function produces outputs with decreasing scale as $\alpha$ increases, due to truncation and zero-padding.
To account for this, we use a relative mean squared error, defined as
\begin{equation}
    \text{Relative MSE} = \frac{\frac{1}{T} \sum_{t=1}^T |\hat{y}(t) - y(t)|^2}{\frac{1}{T} \sum_{t=1}^T |y(t)|^2},
\end{equation}
where $T$ is the sequence length.
This normalizes the loss by the average energy of the true output sequence, making it comparable across different memory functions and values of $\alpha$.
The denominator is always positive by construction, ensuring the metric is well-defined.

\subsection{Experimental Settings}

We now provide complete details of our experimental configuration.
For each memory function and model pair, we conduct multiple experimental runs across different $\alpha$ values, model sizes $m$, and random seeds.
This approach ensures comprehensive coverage of the parameter space, capturing both model sensitivity and robustness across different memory structures.

\paragraph{Architectures Evaluated}

We evaluate four representative sequence modeling architectures spanning different paradigms:
LSTM \cite{hochreiter1997.LongShortTermMemory} (recurrent),
S4D \cite{gu2022.ParameterizationInitializationDiagonal} (structured state-space),
TCN \cite{bai2018.EmpiricalEvaluationGeneric} (convolutional),
and Transformers \cite{vaswani2017.AttentionAllYou} (attention-based).
These architectures are selected based on their theoretical foundations and practical importance in sequence modeling.
We use $m$ to denote the primary model size parameter across all architectures.
For LSTM, $m$ corresponds to the hidden dimension.
For S4D, it is the model dimension ($d_\text{model}$).
For TCN, it is the number of channels.
For Transformers, we vary both the hidden dimension $m$ and the number of attention heads $n_h$,
where the per-head dimension is given by $m/n_h$ following standard implementations.

\paragraph{Data Generation}

The most critical parameter in the data generation process is the sequence length $T$.
The sequence length must be long enough to capture the full range of temporal dependencies, especially for larger values of $\alpha$, to avoid limiting the model's ability to learn long-term patterns.
However, increasing the sequence length also significantly impacts training speed and memory consumption.

To balance these considerations, we set the sequence length to 1024 for LSTM, S4D, and TCN, and to 128 for Transformer models, due to the quadratic complexity of self-attention.
These lengths were selected based on preliminary experiments to ensure adequate coverage of temporal structures without excessive computational overhead.

Other key data generation settings are as follows:
\begin{itemize}
\item \textbf{Numerical precision:} All experiments use double precision (float64) to ensure numerical accuracy.
\item \textbf{Activation functions:} The nonlinear activation functions $\sigma_1$ and $\sigma_2$ in the target generation formula \cref{eq: data generation} are both set to $\tanh$.
\item \textbf{Scaling parameters:} We choose $\alphamax = 0.3$ for $\mupoly$ to ensure that the tail sum of $\rhop$ is sufficiently small (approximately $10^{-8}$ scale). The corresponding values for $\muexp$ are solved numerically to satisfy the condition in \cref{sec: scaling functions}.
\item \textbf{Input distribution:} The input sequences $x(t)$ are generated as i.i.d. random numbers such that $x(t) \sim \mathcal{N}(0,1)$.
This design choice ensures that all temporal dependencies in the output arise solely from the memory function $\rho(s,\alpha)$, not from correlations in the input.
While real-world data often exhibits temporal correlations in both inputs and target mappings, using i.i.d. inputs allows us to isolate and systematically study how different architectures handle the temporal structure encoded in $\rho$. 
This controlled setting is essential for the interpretability of our benchmarking framework, enabling us to precisely attribute observed model behaviors to specific memory properties.
\end{itemize}

\paragraph{Model Parameters}

For each model, we vary the model size to capture the impact of capacity on approximation behavior.
The chosen sizes are designed to be neither too small (preventing learning even for small $\alpha$) nor too large (enabling easy fitting regardless of $\alpha$).
This balance ensures effective measurement of how model size influences approximation performance.

For TCN, we fix the number of layers at 10 across all model sizes to ensure that the receptive field covers the entire sequence length, and vary only the number of channels to investigate its effect on approximation behavior.

For Transformers, we vary both the hidden dimension and the number of attention heads,
as these two parameters together determine the per-head dimensionality,
which plays a critical role in the observed trade-off behaviors.
This dual-parameter variation allows us to investigate how these architectural choices interact with the memory structure and the target's temporal properties.

\paragraph{Training Settings}

To ensure fair comparison, we use consistent training settings across all $\alpha$ values and memory functions.
Each experiment is repeated across multiple random seeds to account for initialization variance and stochastic optimization dynamics.
Specific hyperparameters (optimizer, learning rate, batch size, number of training steps) are chosen based on preliminary experiments and are detailed in the appendix.

We note that since our training data is generated from simple random distributions, gradient estimates remain accurate even with relatively small batch sizes.
We verified this by analyzing the cosine similarity between gradients computed with different batch sizes, confirming stable gradient quality (see appendix for details).

\revision{ 

\paragraph{Length Ablation}
To verify that the observed trends are not driven by the selected sequence lengths, 
we conduct additional experiments at $T \in \{512, 2048\}$ for LSTM, S4D, and TCN, and at $T \in \{64, 256\}$ for Transformer. 
The relative difficulty ordering across memory functions and the sensitivity patterns reported in \cref{sec:results} remain consistent across all tested lengths. This robustness is expected given the scaling methodology described in \cref{subsec: scaling}, where the scaling functions $\mu(\alpha)
$ are defined relative to $T$, ensuring that the temporal structure of each target scales proportionally with sequence length. Full results are provided in \cref{appendix:length_ablation}.} 
\section{Experimental Results and Analysis}\label{sec:results}

In this section,
we present the experimental results by applying the framework to the four sequence modeling architectures described in \cref{sec:methodology}.
We evaluate LSTM and S4D (recurrent and state-space models), TCN (convolutional), Transformer (attention-based), and mixed architectures, systematically analyzing how each handles the four memory functions.
Our analysis serves dual purposes: validating existing theoretical insights regarding these models' ability to capture temporal dependencies, and uncovering new phenomena that have not been theoretically studied.

We organize our findings as follows.
\cref{subsec:recurrent} examines recurrent architectures, confirming their sensitivity to decay rates while revealing how depth impacts performance differently across memory structures.
\cref{subsec:tcn} analyzes TCNs, introducing a novel tail energy complexity measure that better characterizes approximation difficulty beyond binary sparsity.
\cref{subsec:transformer} investigates Transformers through the lens of effective rank theory, revealing critical trade-offs between attention head count and per-head dimensionality.
Finally, \cref{subsec:mixed} demonstrates that sequential composition of architectures can substantially compensate for individual weaknesses.
\cref{tab:alpha_sensitivity} provides a quantitative summary of sensitivity to temporal strength across all architectures, with bold entries highlighting the best-performing architecture for each memory function.

\subsection{Recurrent Architectures}\label{subsec:recurrent}

\begin{table}[h]
\centering
\caption{Sensitivity to temporal strength across all architectures, showing the relative loss increment from minimum to maximum $\alpha$ (geometric mean $\pm$ SEM across model sizes). Bold entries indicate the best-performing architecture for each memory function.}
\label{tab:alpha_sensitivity}
\begin{tabular}{lcccc}
\hline
Model & Exponential & Polynomial & Impulse & Airy \\
\hline
LSTM
& $\mathbf{5.77 \pm 1.50 \times 10^{-1}}$
& $2.01 \pm 1.11 \times 10^{0}$
& $1.25 \pm 0.64 \times 10^{7}$
& $1.39 \pm 7.44 \times 10^{3}$ \\
\hline
S4
& $1.56 \pm 1.36 \times 10^{0}$
& $\mathbf{1.16 \pm 1.45 \times 10^{0}}$
& $1.04 \pm 1.72 \times 10^{5}$
& $1.85 \pm 1.29 \times 10^{1}$ \\
\hline
TCN
& $3.14 \pm 1.08 \times 10^{1}$
& $4.26 \pm 0.61 \times 10^{1}$
& $\mathbf{9.15 \pm 6.01 \times 10^{-1}}$
& $\mathbf{1.01 \pm 0.19 \times 10^{1}}$ \\
\hline
Transformer
& $1.22 \pm 0.51 \times 10^{1}$
& $2.71 \pm 1.44 \times 10^{1}$
& $4.09 \pm 6.21 \times 10^{2}$
& $8.13 \pm 1.49 \times 10^{1}$ \\
\hline
\end{tabular}
\end{table}

\begin{figure}[!ht]
    \centering
    \includegraphics[width=0.85\linewidth]{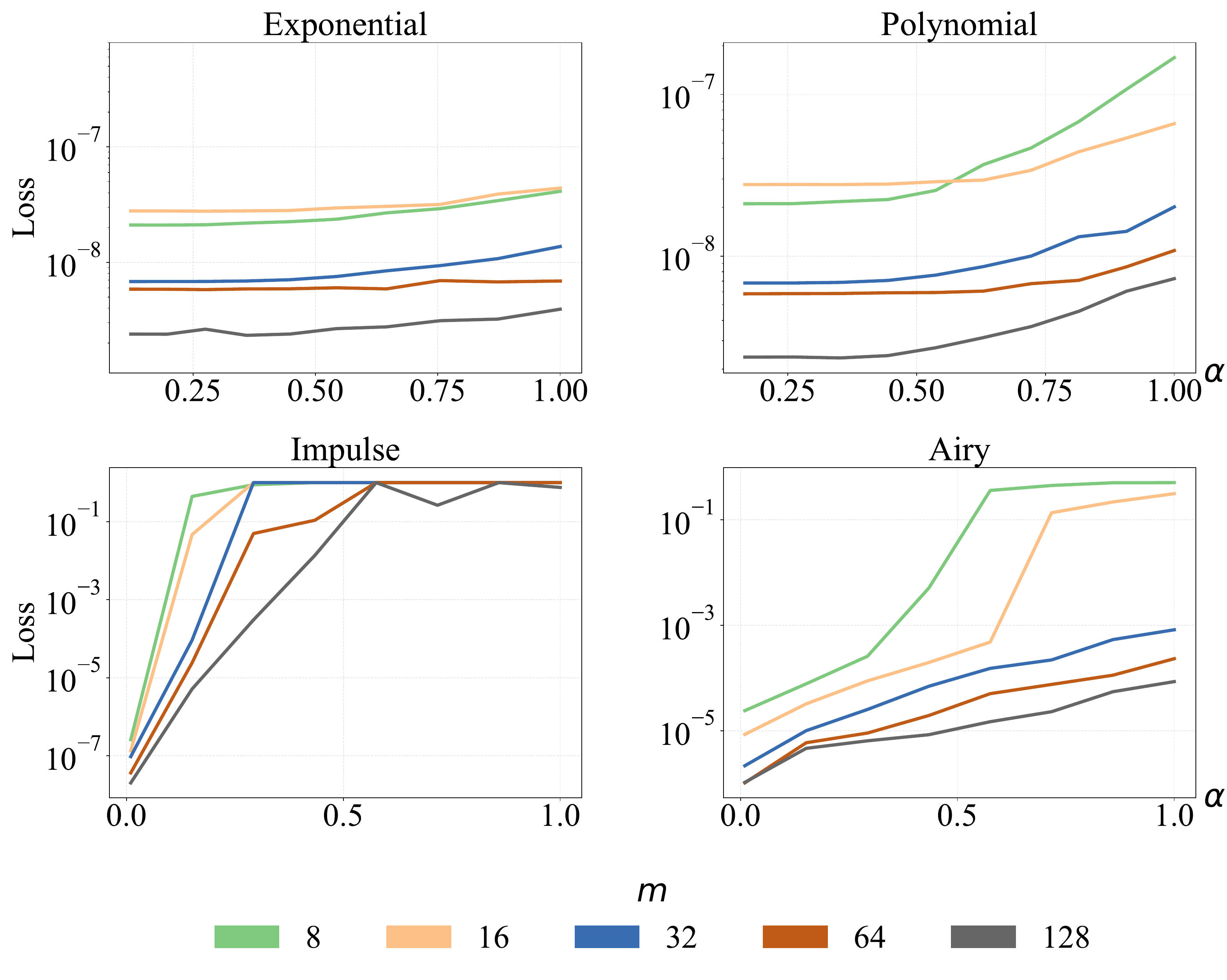}
    \caption{Loss versus $\alpha$ for LSTM on all four memory functions, with varying model sizes $m$.}
    \label{fig: lstm}
\end{figure}

\begin{figure}[!ht]
    \centering
    \includegraphics[width=0.85\linewidth]{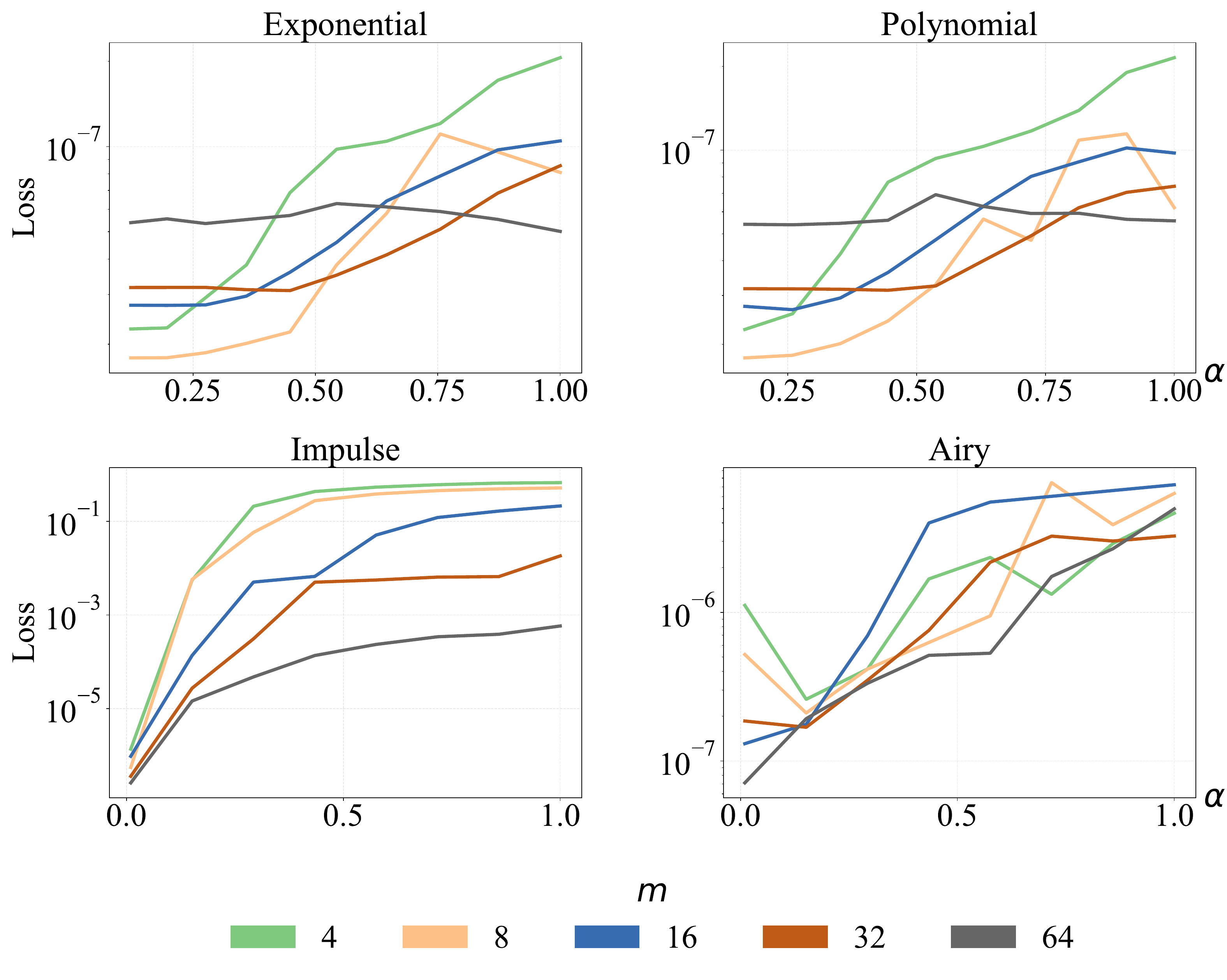}
    \caption{Loss versus $\alpha$ for S4D on all four memory functions, with varying model sizes $m$.}
    \label{fig: s4d}
\end{figure}

\begin{observation}\label{obs:recurrent_decay}
Recurrent architectures (LSTM and S4D) are strongly influenced by memory decay rates and dependency range, performing well on exponential decay but degrading under polynomial decay and sparse long-range dependencies.
\end{observation}

As shown in \cref{fig: lstm,fig: s4d} and quantified in \cref{tab:alpha_sensitivity} (rows 1--2),
both LSTM and S4D maintain low and stable loss across varying values of $\alpha$ under the exponential memory function $\rhoe$.
This indicates that these architectures are well-suited for approximating sequences with rapidly decaying temporal dependencies.
This behavior is supported by the theoretical analysis in \cite{miller2019.StableRecurrentModels,li2022.ApproximationOptimizationTheory,wang2023.InverseApproximationTheory},
which demonstrates that recurrent architectures can efficiently approximate targets with exponential memory decay.
Additionally, \cite{wang2023.StatespaceModelsLayerwise} shows that state-space models with layer-wise nonlinearity also adapt naturally to exponential memory structures.

However, under the polynomial memory function $\rhop$,
both models exhibit a noticeable increase in loss as $\alpha$ increases.
This suggests a significant limitation when approximating slowly decaying temporal structures,
where past inputs retain a non-negligible influence over long timescales.
This observation is consistent with the findings of \cite{zhao2020.RNNLSTMHave},
which shows that RNNs and LSTMs struggle to capture polynomial decaying dependencies without explicit architectural modifications.
Similarly, \cite{li2022.ApproximationOptimizationTheory} demonstrates that linear recurrent models face fundamental approximation limitations under slowly decaying memory.
\cite{wang2024.StableSSMAlleviatingCurse} further argues that standard state-space models also suffer from this limitation by default,
and proposes a stable reparameterization to improve their long-range dependency handling.

When evaluated on the impulse memory function $\rhoi$,
both LSTM and S4D show a rapid increase in loss as $\alpha$ grows.
This highlights a severe limitation in modeling sparse, long-range dependencies,
where the target depends on a single distant input.
This difficulty is investigated in
\cite{bengio1994.LearningLongtermDependencies,li2022.ApproximationOptimizationTheory,wang2023.InverseApproximationTheory},
which shows that recurrent architectures struggle to approximate sparse, isolated dependencies.
Empirical evidence supporting this limitation is also provided in \cite{bai2018.EmpiricalEvaluationGeneric}.

The Airy memory function $\rho_{\text{Ai}}$ exhibits a similar degradation pattern to the polynomial function,
with loss increasing as $\alpha$ grows.
This is consistent with the fact that both functions involve non-sparse, distributed temporal dependencies that become more challenging to capture as the temporal strength increases.
 
\revision{
\paragraph{Explanation via Prony series}
The contrast between exponential and polynomial difficulty admits a unified analytical explanation via the Prony series perspective.
The key observation is that the polynomial memory function $\rhop$ can itself be approximated by a Prony series, which is a sum of exponentials with modes spread across many timescales.
Since recurrent architectures implement Prony series through their hidden state dynamics,
fitting $\rhop$ requires the optimizer to simultaneously find exponential modes spanning many orders of magnitude,
a task that becomes increasingly ill-conditioned as the spread grows.
To verify this directly, we define a $K$-mode Prony series target with controllable spread parameter $\gamma$:
\begin{equation}\label{eq:prony_target}
    \rho_{K,\gamma}(s) = \frac{1}{K}\sum_{k=1}^{K} \lambda_k^s,
    \quad \lambda_k = \exp\!\left(-\frac{10^{\gamma(k-1)/(K-1)}}{T}\right),
\end{equation}
where $\gamma = 0$ concentrates all modes at a single timescale (effectively a single exponential),
and increasing $\gamma$ spreads modes across more orders of magnitude, approaching the behavior of $\rhop$ in the large-$\gamma$ limit.

\cref{fig:prony} shows LSTM loss as a function of $\gamma$ (left, $K=8$ fixed) and $K$ (right, $\gamma=0.5$ fixed).
Loss increases with $\gamma$ for smaller models but remains stable for larger models,
while no systematic trend with $K$ is observed at any model size.
This confirms that it is the \emph{spread} of decay rates across timescales, not their count, 
that determines recurrent architectures' approximation difficulty.
Exponential decay ($\gamma \approx 0$) concentrates all modes at a single timescale and is easy to match, 
while polynomial decay, viewed as a large-$\gamma$ Prony series, requires modes spread across many orders of magnitude,
creating an ill-conditioned optimization landscape that difficult for recurrent models to learn.

\begin{figure}[!ht]
    \centering
    \includegraphics[width=0.85\linewidth]{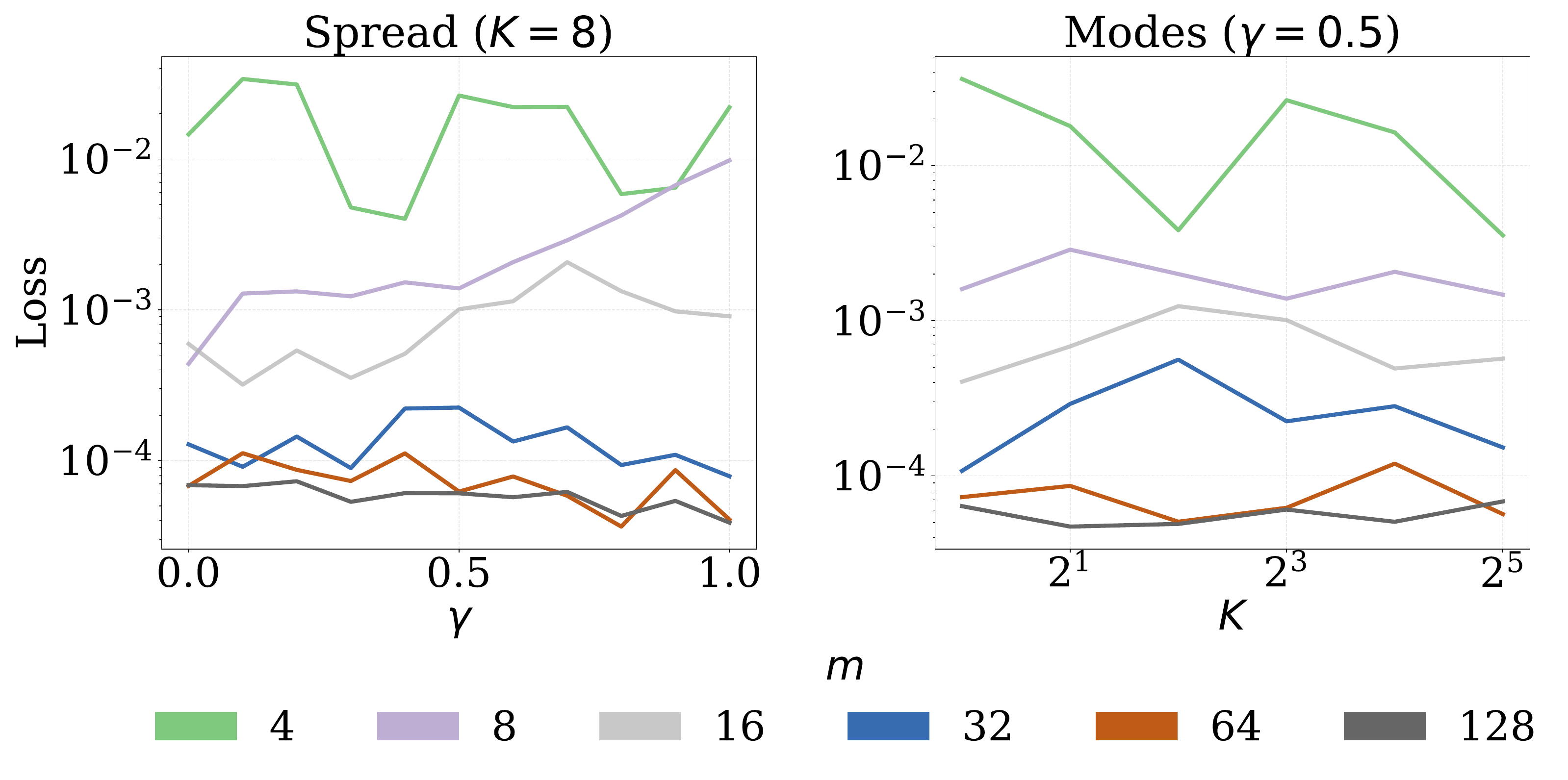}
    \caption{LSTM loss on the $K$-mode Prony series target $\rho_{K,\gamma}$ \cref{eq:prony_target},
    varying spread $\gamma$ with $K=8$ fixed (left) and varying mode count $K$ with $\gamma=0.5$ fixed (right),
    across model sizes $m$.
    Increasing $\gamma$ leads to higher loss, confirming that the spread of decay rates is the primary source of difficulty for recurrent architectures.
    Loss remains stable across values of $K$, indicating that mode count does not contribute to approximation difficulty.} 
    \label{fig:prony}
\end{figure}
}

\begin{figure}[!ht]
    \centering
    \includegraphics[width=0.85\linewidth]{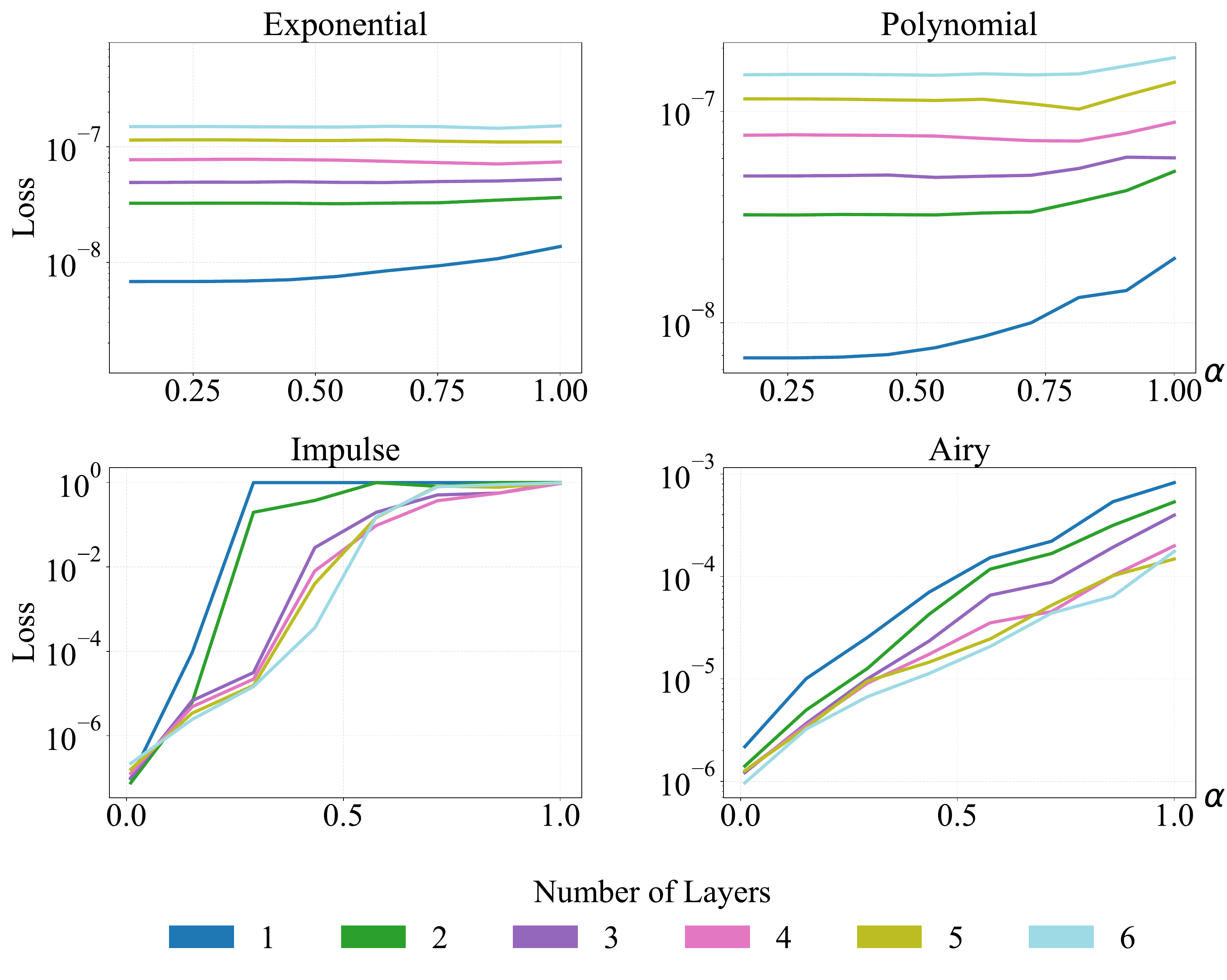}
    \caption{Loss versus $\alpha$ for multilayer LSTM on all four memory functions, with fixed $m=32$ and varying number of layers.}
    \label{fig: multilayer_lstm}
\end{figure}

\begin{figure}[!ht]
    \centering
    \includegraphics[width=0.85\linewidth]{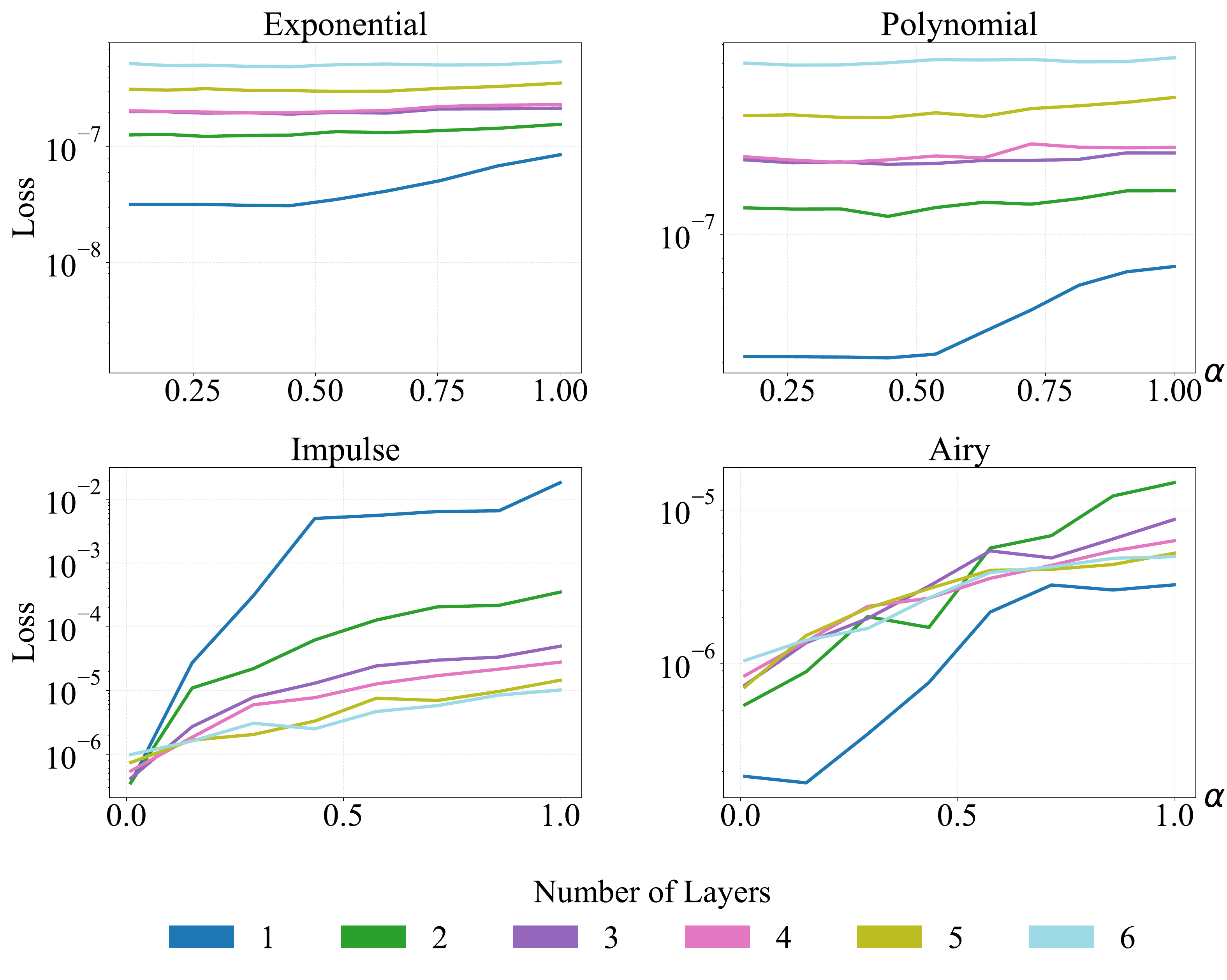}
    \caption{Loss versus $\alpha$ for multilayer S4D on all four memory functions, with fixed $m=32$ and varying number of layers.}
    \label{fig: multilayer_s4d}
\end{figure}

Given the distinct sensitivity patterns observed across memory functions, we investigate how depth affects recurrent architectures by fixing $m=32$ and varying layers from 1 to 6 (\cref{fig: multilayer_lstm,fig: multilayer_s4d}).

\begin{observation}\label{obs:recurrent_depth}
Depth effects in recurrent architectures depend critically on memory function structure: for $\rhoe$ and $\rhop$, a single layer is sufficient, whereas sparse long-range dependencies ($\rhoi$, $\rho_{\text{Ai}}$) require 3+ layers to avoid severe degradation at large $\alpha$.
\end{observation}

As shown in \cref{fig: multilayer_lstm,fig: multilayer_s4d},
for the exponential and polynomial memory functions,
single-layer networks consistently achieve the lowest loss across all $\alpha$ values,
with performance progressively degrading as additional layers are added.
The loss curves remain flat across $\alpha$ values for all depths, indicating that depth does not alter sensitivity to temporal strength.
For smoothly-decaying dependencies, additional layers provide no benefit and may harm optimization.
The single-layer recurrent cell appears sufficient to capture exponential and polynomial decay patterns,
consistent with theoretical analyses showing that these architectures naturally align with exponentially decaying memory structures \cite{miller2019.StableRecurrentModels,wang2023.StatespaceModelsLayerwise}.

In contrast, for the impulse and Airy memory functions,
depth provides substantial benefits.
Single-layer networks exhibit the most severe performance degradation as $\alpha$ increases,
particularly for the impulse function where the loss increases by several orders of magnitude.
Two-layer networks also struggle significantly on these memory structures.
Networks with 3+ layers maintain significantly better approximation quality, with converging loss curves.
Multi-layer architectures benefit targets with longer dependency ranges: if one layer captures range $l$, then $k$ stacked layers can together capture range $kl$.
This multiplicative effect is further investigated in \cite{bao2025.EffectDepthExpressivity} for linear deep state-space models.
This observation aligns with empirical findings in \cite{bai2018.EmpiricalEvaluationGeneric},
which demonstrate that deeper architectures show improved performance on tasks requiring long-range pattern recognition.

Beyond approximation quality,
we also investigate the impact of memory structure on training speed by measuring the number of training steps required to reach a specified error threshold as a function of $\alpha$.

\begin{figure}[!ht]
    \centering
    \includegraphics[width=0.85\linewidth]{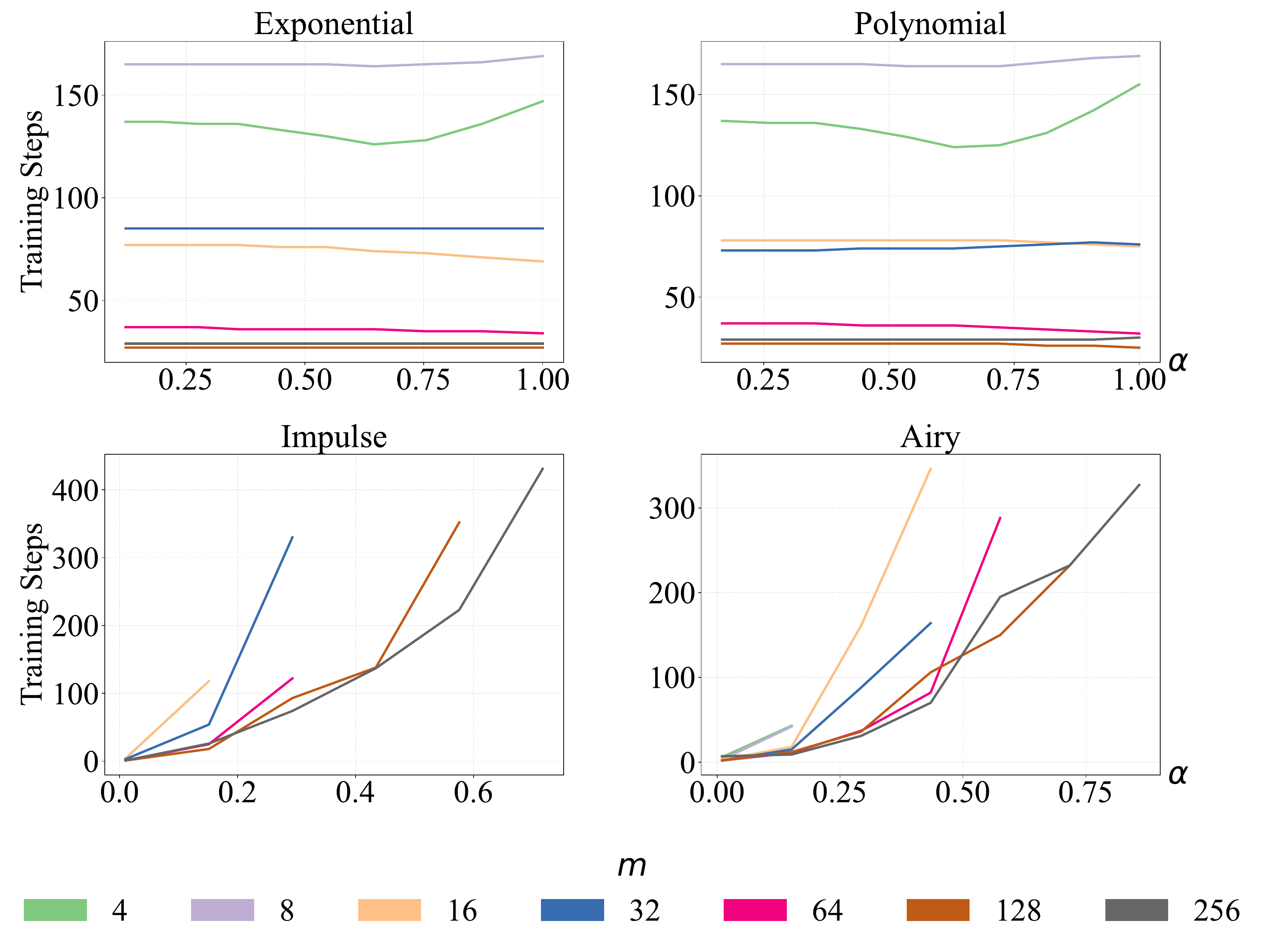}
    \caption{Training speed for LSTM on exponential, polynomial, and impulse memory functions, showing the number of training steps required to reach a specified error threshold versus $\alpha$.}
    \label{fig: LSTM optimization}
\end{figure}

As shown in \cref{fig: LSTM optimization},
for both the exponential and polynomial memory functions,
the number of training steps required to reach the error threshold remains relatively stable across different $\alpha$ values.
In contrast,
the number of training steps for the impulse memory function $\rhoi$ increases dramatically as $\alpha$ grows.
This phenomenon is theoretically supported by \cite{li2022.ApproximationOptimizationTheory},
which demonstrates that the training time of linear recurrent models scales exponentially with the dependency range.
This analysis provides a foundation for comparing training dynamics across different architectures,
which we explore further in the TCN section below.

\subsection{Temporal Convolutional Networks}\label{subsec:tcn}

\begin{figure}[!ht]
    \centering
    \includegraphics[width=0.85\linewidth]{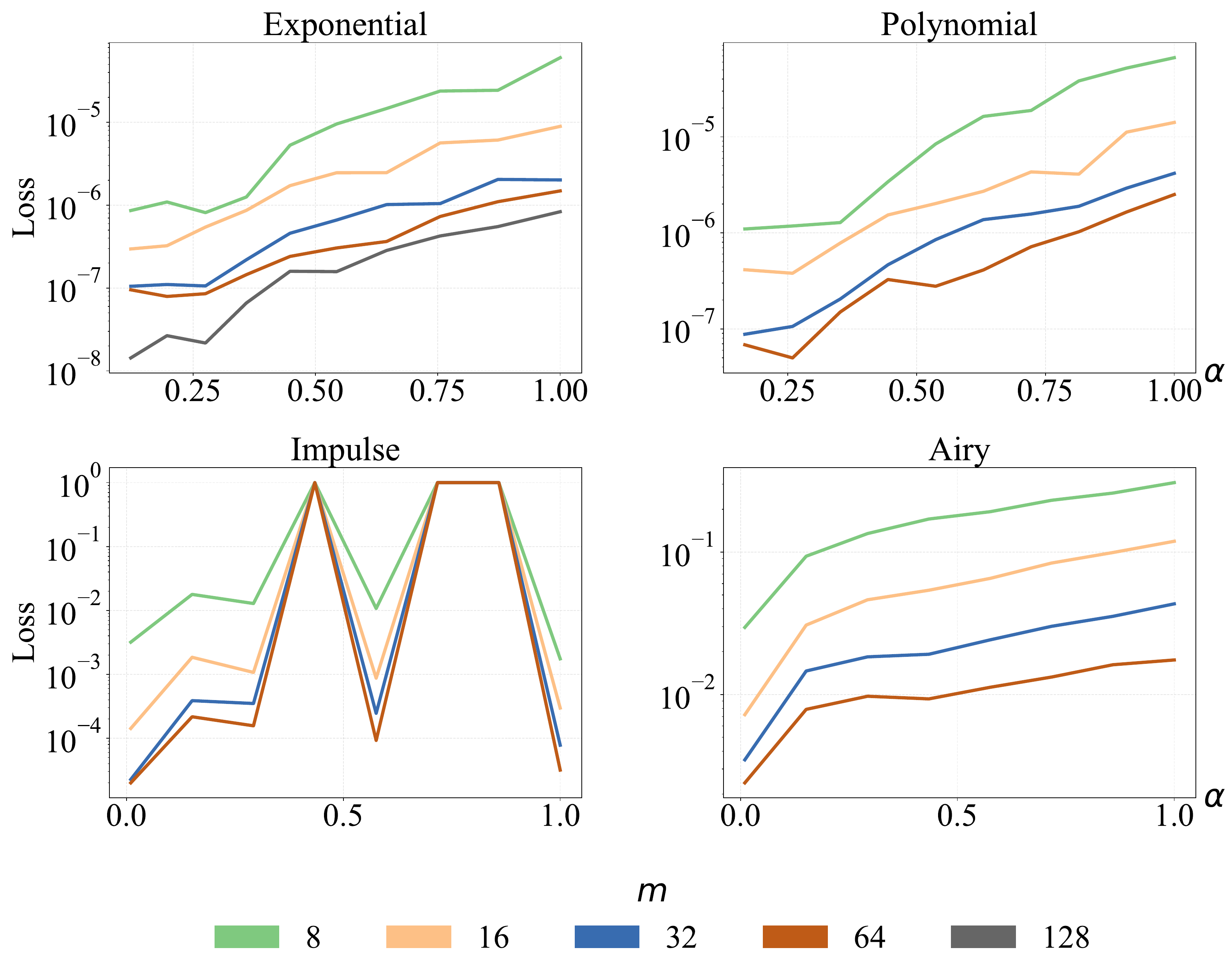}
    \caption{Loss versus $\alpha$ for TCN on all four memory functions, with varying model sizes $m$.}
    \label{fig: tcn}
\end{figure}

\begin{observation}\label{obs:tcn_impulse}
The approximation behavior of TCN is not affected by the dependency range of $\rhoi$.
\end{observation}

In \cref{fig: tcn},
TCNs maintain stable performance across varying $\alpha$ under the impulse memory function $\rhoi$,
where the output depends on a single input at a distant time step.
This demonstrates TCNs' ability to capture long-range dependencies through dilated convolutions,
which expands the receptive field exponentially with network depth.
This observation is also supported by theoretical results in \cite{jiang2021.ApproximationTheoryConvolutional},
which shows that TCNs efficiently approximate functions with sparse or distant input-output dependencies.
The empirical trend confirms this prediction, with observed fluctuations attributable to optimization challenges rather than approximation limitations, as we explore in our analysis of initialization sensitivity and training speed below.

\begin{observation}\label{obs:tcn_sparsity}
The approximation behavior of TCNs cannot be solely characterized by sparsity.
\end{observation}

As shown in \cref{fig: tcn},
TCNs exhibit increased approximation error under the Airy memory function $\rho_{\text{Ai}}$.
In this case,
larger values of $\alpha$ reduce the sparsity,
making the target function more challenging to approximate.
In contrast, the impulse memory function $\rhoi$ remains maximally sparse regardless of $\alpha$,
leading to stable approximation performance.
This aligns with theoretical insights from \cite{jiang2021.ApproximationTheoryConvolutional},
which shows that TCNs are more effective at approximating targets with higher sparsity,
where each channel can be utilized to represent a single nonzero value of the memory function.

However, as shown in \cref{fig: tcn},
the performance of TCNs also degrades as $\alpha$ increases under both exponential and polynomial memory functions.
This behavior cannot be fully explained by conventional binary sparsity measures,
since both $\rhoe$ and $\rhop$ maintain fixed sparsity regardless of $\alpha$,
which would predict constant error across different $\alpha$ values.
This suggests that a more precise notion of complexity is required to capture the observed trend.

To address this gap, we introduce a novel \textbf{tail energy complexity measure} that captures magnitude decay of memory functions beyond binary sparsity.
We define this measure as
\begin{align}\label{eq: complexity measure}
    C(\rho, s) := \sum_{t=s}^T \abs{\rho(\pi(t), \alpha)}^2,
\end{align}
where the function $\pi : \{0, \ldots, T\} \rightarrow \{0, \ldots, T\}$
is a permutation that reorders the values of the memory function
so that the composed function $\rho \circ \pi$ is monotone non-increasing.
This measure captures the tail of the memory function after accounting for the largest values.

\begin{figure}[!ht]
    \centering
    \includegraphics[width=0.85\linewidth]{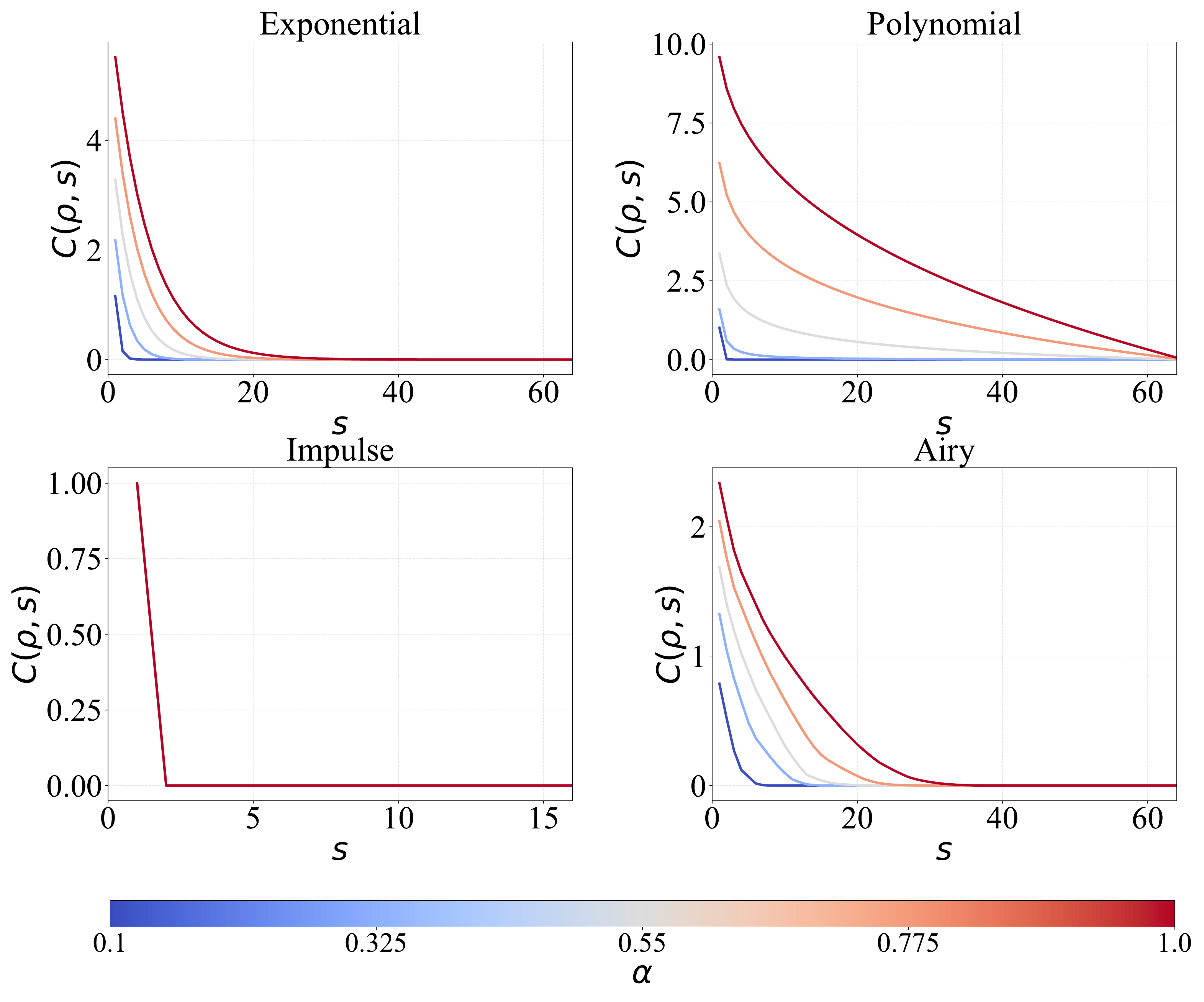}
    \caption{Complexity measures $C(\rho, s)$ and their correlation with training loss for TCN across all four memory functions. The top row shows the complexity measure versus $s$ for different $\alpha$ values, while the bottom row shows the relationship between loss and complexity for model size $m=64$. Comparing with \cref{fig: tcn}, we observe that for $\rhoe$, $\rhop$, and $\rho_{\text{Ai}}$, both complexity and loss increase with $\alpha$, and the bottom row confirms a positive correlation between loss and complexity, indicating that complexity explains the loss trends observed in \cref{fig: tcn}. For $\rhoi$, the complexity measure is constant zero, consistent with the stable performance across $\alpha$ values shown in \cref{fig: tcn}.}
    \label{fig: complexity plot}
\end{figure}

In \cref{fig: complexity plot},
we plot this complexity measure for each memory function.
The results show a clear positive correlation between the complexity measure and the observed loss for $\rhoe$, $\rhop$, and $\rho_{\text{Ai}}$,
indicating that this measure can effectively capture the increasing approximation difficulty for these memory structures.
In contrast, the complexity measure for $\rhoi$ is constant zero,
reflecting the inherently sparse nature of the impulse function,
which aligns with the stable loss observed in this case.
This suggests that the proposed complexity measure provides a meaningful estimate of the approximation bound for TCNs.
While this measure may not yield bounds as tight as the rank-based approach in \cite{jiang2021.ApproximationTheoryConvolutional}, it is straightforward to compute and provides a practical characterization of TCN approximation power.
We formalize this intuition in \cref{thm: tcn approximation} (appendix), demonstrating that $C(\rho, s)$ provides a concrete upper bound on TCN approximation error.
The theorem extends \cite{jiang2021.ApproximationTheoryConvolutional} by incorporating tail energy complexity, directly linking empirical observations to theoretical guarantees.

\begin{observation}\label{obs:tcn_optimization}
The optimization behavior of TCNs is influenced by both the target memory function and the choice of optimizer.
\end{observation}

\begin{figure}[!ht]
    \centering
    \includegraphics[width=0.85\linewidth]{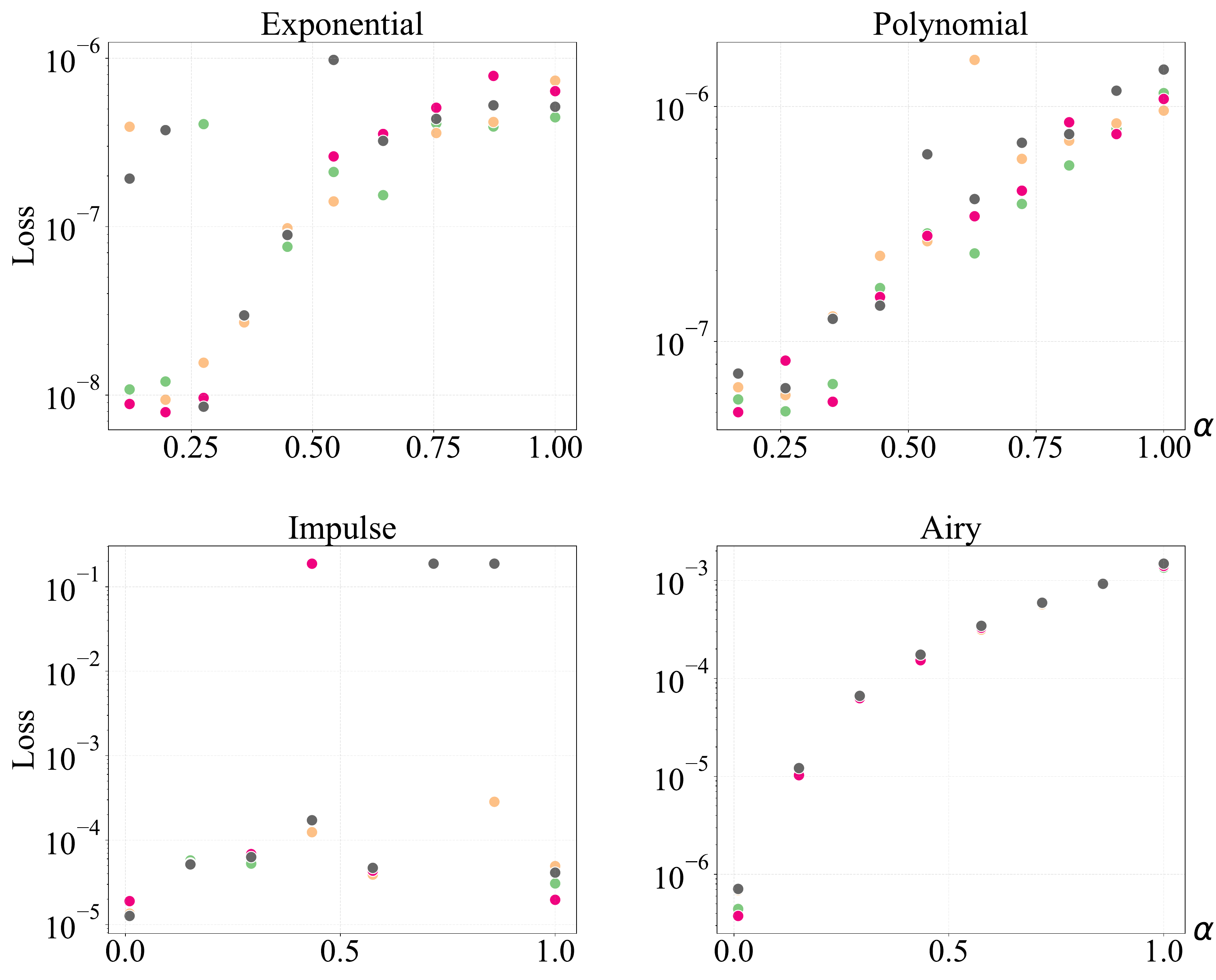}
    \caption{TCN training loss variability across different random seeds using the Adam optimizer, showing final loss distributions for all four memory functions at different $\alpha$ values.}
    \label{fig: tcn_adam}
\end{figure}

\begin{figure}[!ht]
    \centering
    \includegraphics[width=0.85\linewidth]{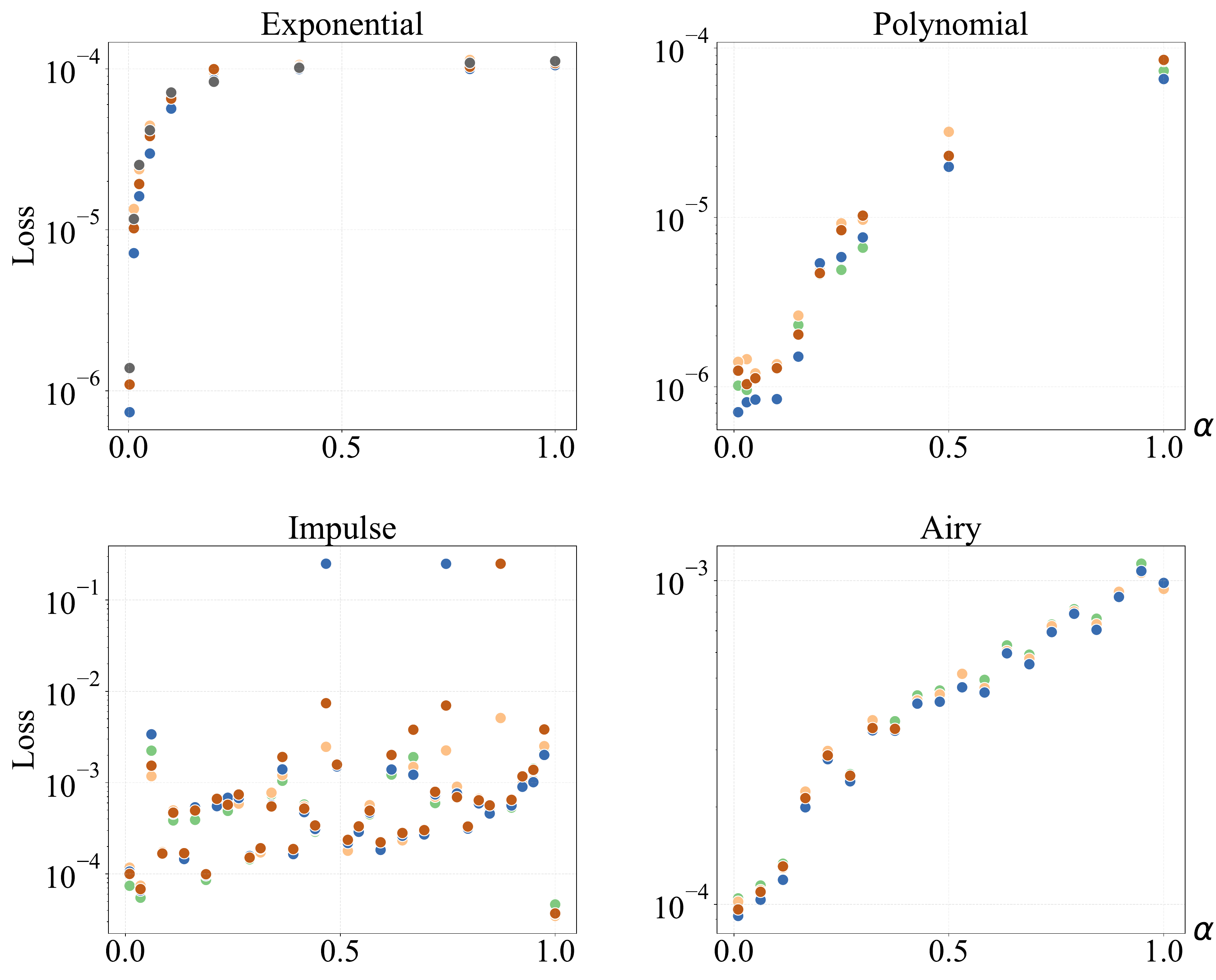}
    \caption{TCN training loss variability across different random seeds using the AdamW optimizer, showing final loss distributions for all four memory functions at different $\alpha$ values.}
    \label{fig: tcn_adamW}
\end{figure}

Both the memory function and optimizer choice significantly impact the training stability of TCNs.
As shown in \cref{fig: tcn_adam,fig: tcn_adamW},
TCNs trained with the Adam optimizer exhibit relatively stable loss across different seeds for exponential and polynomial memory functions,
indicating more consistent convergence.
In contrast, the impulse memory function leads to significantly more variable training outcomes,
with some seeds achieving much lower loss than others.

When using the AdamW optimizer, this variability becomes even more pronounced,
especially for the impulse and Airy memory functions,
where the loss spread is much wider.
Overall, these results indicate that TCNs are more sensitive to initialization than other architectures,
and that this sensitivity depends on both the memory function and the optimizer choice.

To complement our analysis of initialization sensitivity,
we also examine the training speed of TCNs in relation to memory structure and model capacity.

\begin{figure}[!ht]
    \centering
    \includegraphics[width=0.85\linewidth]{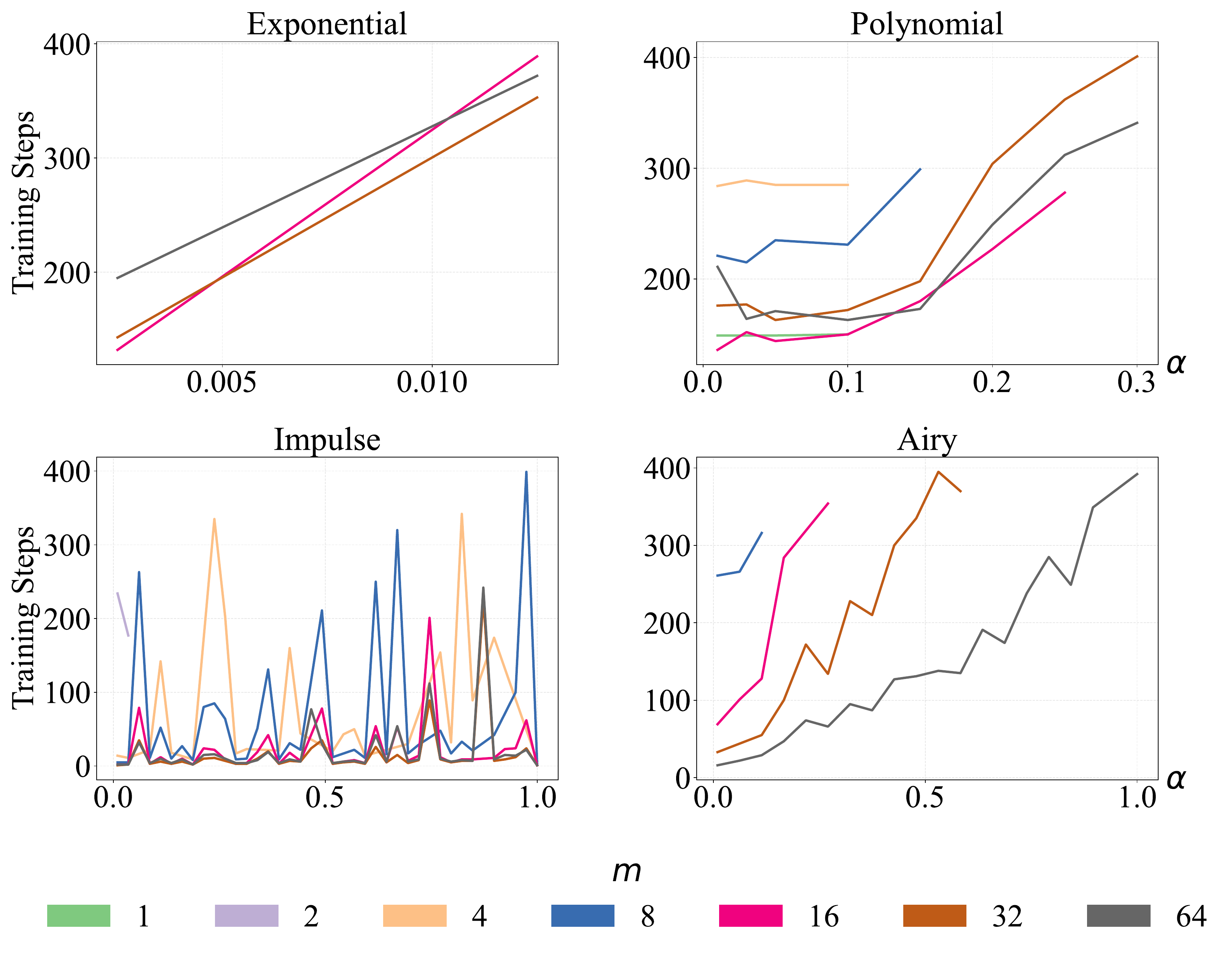}
    \caption{Training speed for TCN on exponential, polynomial, and impulse memory functions, showing the number of training steps required to reach a specified error threshold versus $\alpha$ for different model sizes.}
    \label{fig: TCN optimization}
\end{figure}

As shown in \cref{fig: TCN optimization},
when the number of channels is sufficiently large,
the training speed for both exponential and polynomial memory functions remains relatively stable across different $\alpha$ values,
similar to the pattern observed for LSTM in \cref{fig: LSTM optimization}.
However,
when the number of channels is insufficient,
the required training steps increase significantly,
indicating that model capacity plays a crucial role in training efficiency.
For the impulse memory function,
the training dynamics are notably unstable,
further confirming our previous observation that TCN training is sensitive to initialization.
These results demonstrate that both the memory structure and the model capacity significantly influence the training speed of TCNs,
highlighting the interplay between approximation power and optimization dynamics.
Having examined convolutional approaches, we now turn to attention-based architectures, where the effective rank framework provides theoretical insights into how Transformers balance expressiveness and approximation capacity.

\subsection{Transformers}\label{subsec:transformer}

\begin{figure}[!ht]
    \centering
    \includegraphics[width=0.85\linewidth]{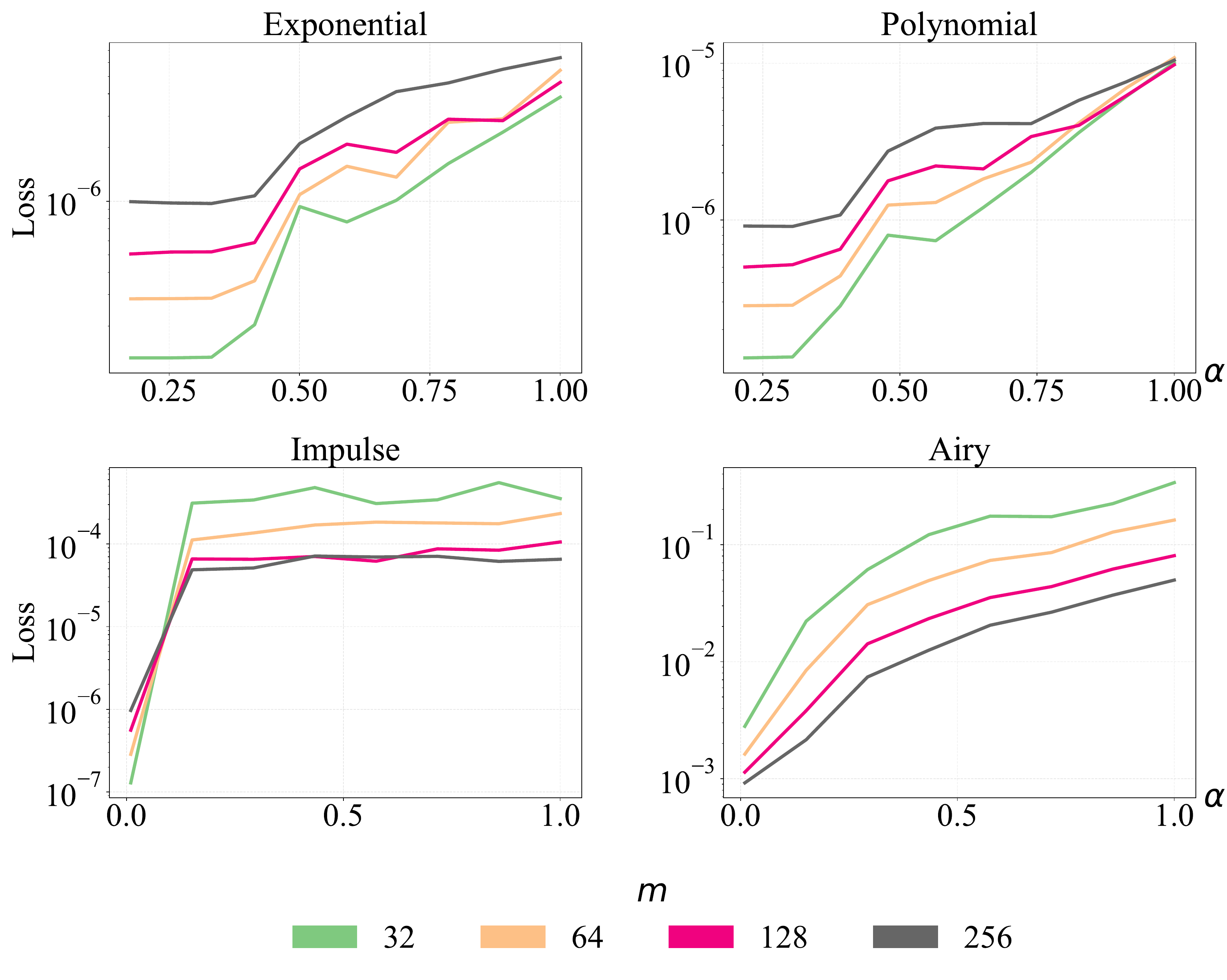}
    \caption{Loss versus $\alpha$ for Transformer on all four memory functions, with fixed number of heads $n_h=4$ and varying model sizes $m$.}
    \label{fig: trans}
\end{figure}

\begin{observation}\label{obs:transformer_similarity}
The approximation behavior of Transformers exhibits similar trends to TCNs across all four memory functions,
despite not operating as direct convolutions.
\end{observation}

As shown in \cref{fig: trans},
Transformers demonstrate stable performance on the impulse memory function $\rhoi$,
similar to the behavior observed for TCNs.
Additionally,
they exhibit increasing approximation error under the exponential and polynomial memory functions as $\alpha$ increases,
mirroring the patterns seen with TCNs.
This similarity might initially suggest that Transformers function as convolutions,
as explored in \cite{cordonnier2020.RelationshipSelfAttentionConvolutional}.
However,
in our experiments with only 4 attention heads,
the Transformer cannot be directly interpreted as implementing dilated convolutions in the same manner as TCNs.

Instead,
we interpret the observed behavior through the lens of effective rank theory,
as analyzed in \cite{jiang2024.ApproximationRateTransformer}.
This theory demonstrates that the approximation power of Transformers is governed by the rank structure of the target function.
For the exponential, polynomial, and Airy memory functions,
the effective rank increases with $\alpha$,
leading to higher approximation error.
In contrast,
the impulse memory function maintains a constant low rank regardless of $\alpha$,
resulting in stable approximation performance.

\begin{figure}[!ht]
    \centering
    \includegraphics[width=0.85\linewidth]{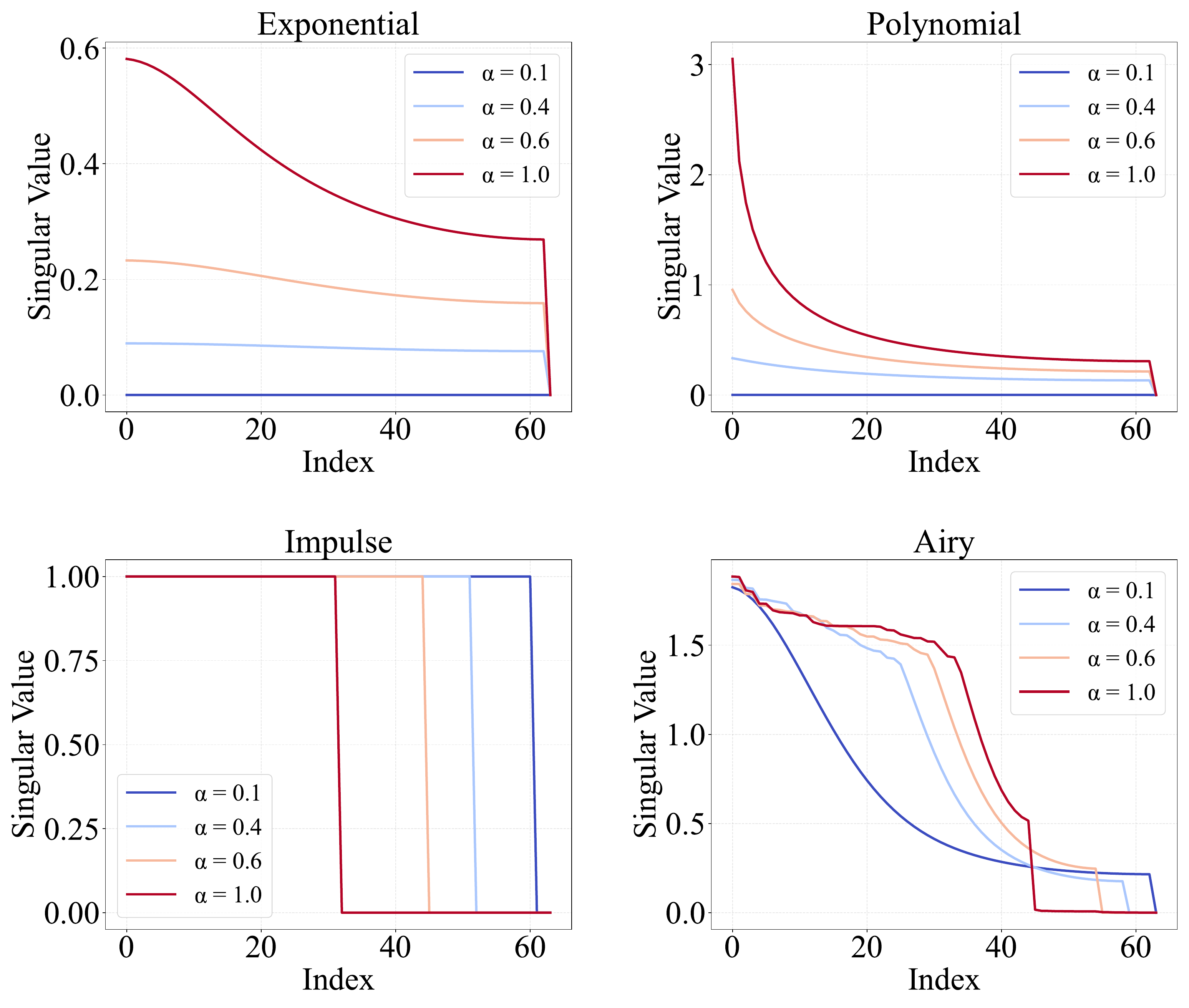}
    \caption{Singular value spectrum of the target memory functions, illustrating the effective rank structure that governs Transformer approximation behavior.}
    \label{fig: singular value plot}
\end{figure}

To validate this rank-based interpretation,
we visualize the singular value spectrum of the target memory functions in \cref{fig: singular value plot}.
Specifically, following the tensorization framework in \cite{jiang2024.ApproximationRateTransformer}, we tensorize each memory function $\rho(s, \alpha)$ by reshaping the sequence $(\rho(0, \alpha), \rho(1, \alpha), \ldots, \rho(T-1, \alpha))$ into a $d$-dimensional tensor $\mathcal{T} \in \mathbb{R}^{n \times n \times \cdots \times n}$ (where $T = n^d$), and compute the singular values of the unfolding matrices to characterize the tensor rank structure.
The results clearly show that the rate of singular value decay varies significantly across memory functions and $\alpha$ values.
For $\rhoe$, $\rhop$, and $\rho_{\text{Ai}}$,
larger values of $\alpha$ lead to slower decay in singular values,
indicating higher effective rank and thus greater approximation difficulty.
For $\rhoi$,
the singular values decay rapidly and consistently across all $\alpha$ values,
confirming the constant low-rank structure that enables stable Transformer performance.
This alignment between the singular value analysis and the observed loss patterns provides strong empirical support for the effective rank perspective on Transformer approximation.

We next investigate the trade-off between number of attention heads $n_h$ and head dimension.
With head dimension set to $m / n_h$ (where $m$ is total model size), increasing heads reduces per-head dimension, creating a fundamental trade-off.

\begin{figure}[!ht]
    \centering
    \includegraphics[width=0.85\linewidth]{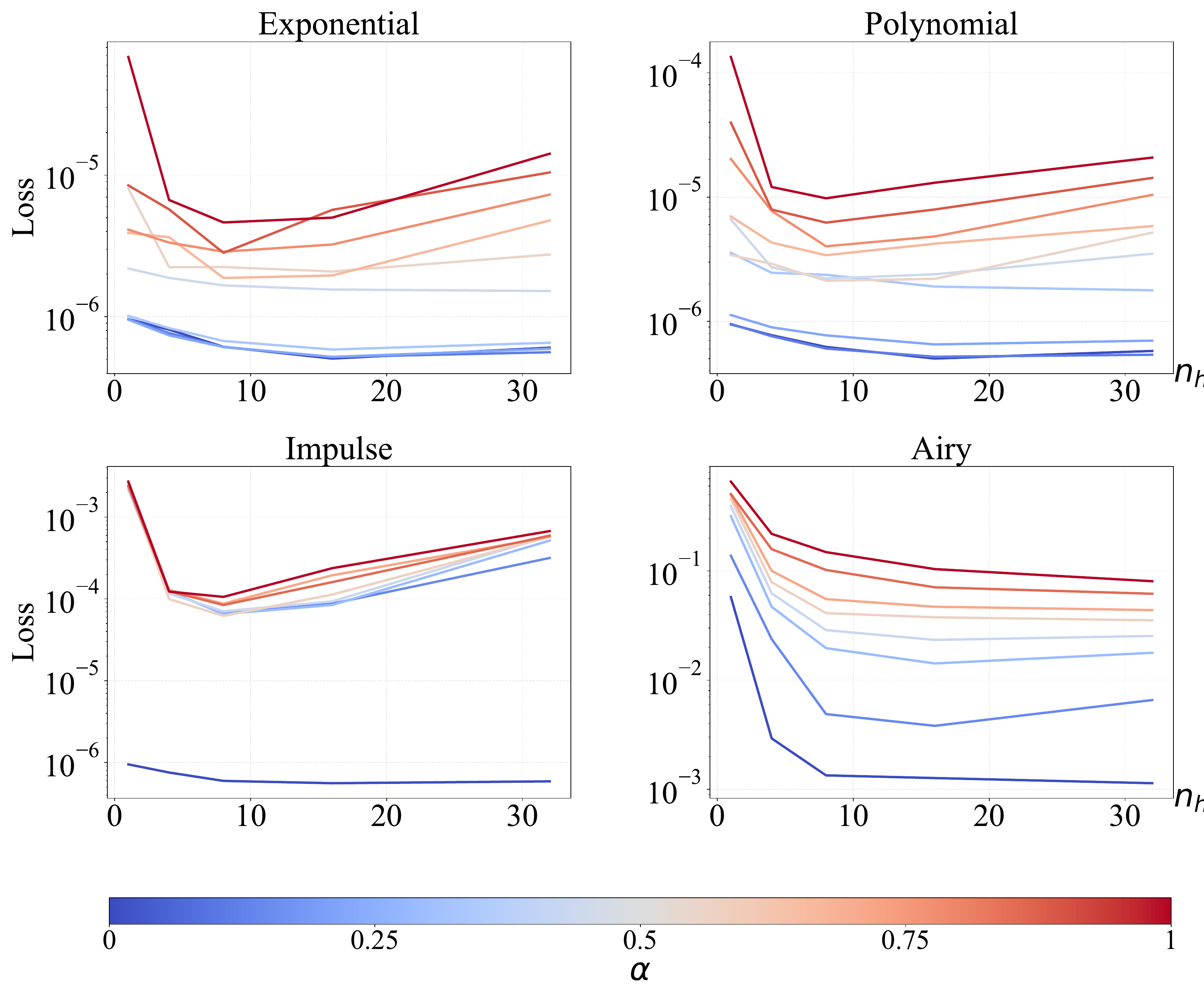}
    \caption{Loss versus number of attention heads $n_h$ for Transformer on all four memory functions, with fixed model size $m=128$ and varying $\alpha$ values.}
    \label{fig: trans tradeoff}
\end{figure}

\begin{observation}\label{obs:transformer_tradeoff}
The optimal number of attention heads depends on both the memory function type and temporal strength $\alpha$, with distinct sensitivity patterns across different memory functions.
\end{observation}

As shown in \cref{fig: trans tradeoff},
the relationship between loss and the number of heads varies significantly across memory functions.
For the exponential and polynomial memory functions at small $\alpha$ values,
the loss decreases monotonically as the number of heads increases,
suggesting that more heads are beneficial when the temporal dependencies are relatively weak.
However,
at larger $\alpha$ values,
the loss curves become non-monotonic,
exhibiting a turning point where additional heads begin to degrade performance.
This suggests that when temporal dependencies are strong,
there exists an optimal balance between the number of heads and the dimension per head.
Similar trade-offs have been observed in \cite{yu2026.EffectAttentionHead} for single-layer Transformers on retrieval-type tasks. 

The impulse memory function exhibits non-monotonic behavior across all $\alpha$ values,
with an optimal number of heads that varies depending on $\alpha$.
This reflects the sparse nature of the impulse function,
where the model must capture a single distant dependency,
and the optimal architecture depends critically on the specific temporal strength.

In contrast,
the Airy memory function shows relatively monotonic behavior,
with loss generally decreasing as the number of heads increases.
This indicates less sensitivity to the head count versus dimension trade-off,
possibly due to the distributed nature of the Airy function's temporal structure.

The optimal Transformer configuration depends critically on the temporal dependency structure, with different memory functions requiring distinct head count and dimension choices.

While each architecture exhibits distinct strengths and weaknesses across different memory functions, we next explore whether sequential composition of complementary architectures can overcome individual limitations.

\subsection{Mixture of Models}\label{subsec:mixed}

Sequential composition, or hybrid architectures, refers to combining different model architectures in a pipeline where the output of one model serves as input to another.
This approach has been successfully applied across various domains: the Vision Transformer (ViT) \cite{dosovitskiy2020.ImageWorth16x16} uses convolutional patch embedding before transformer layers, while architectures like CoAtNet \cite{dai2021.CoAtNetMarryingConvolution} (computer vision) and Conformer \cite{gulati2020.ConformerConvolutionaugmentedTransformer} (speech recognition) combine convolution and attention to leverage their complementary strengths.
Such hybrid designs show potential improvements over individual components.
In the context of sequence modeling with memory functions, we investigate whether similar compositional strategies can help architectures compensate for their individual limitations on specific temporal structures.

\begin{observation}\label{obs:mixed_composition}
Sequential composition of architectures substantially improves performance on memory functions where one architecture performs strongly and another performs poorly, with the mixed model approaching the stronger component's performance.
\end{observation}

We evaluate mixed models where one architecture's output serves as input to another, testing all pairwise combinations of LSTM, S4, TCN, and Transformer.
\cref{tab:mixed_alpha_sensitivity_exp_colored} presents the sensitivity to temporal strength (relative loss increment from minimum to maximum $\alpha$, as defined in \cref{para:alpha_sensitivity}) for the exponential memory function, with additional results for polynomial, impulse, and Airy functions provided in Appendix~\ref{appendix:mixed_tables}.
Diagonal entries represent single-architecture baselines.

Sequential composition exhibits strong synergies when architectures have complementary strengths, particularly when one component is LSTM.
For the exponential memory function, LSTM$\rightarrow$S4 ($0.17$) outperforms both standalone architectures (LSTM: $13.2$, S4: $0.22$).
Similarly, for the polynomial function, LSTM$\rightarrow$S4 ($0.17$) substantially improves over both baselines (LSTM: $2.48$, S4: $0.20$).
Most dramatically, for the impulse function, LSTM$\rightarrow$TCN reduces sensitivity by five orders of magnitude---from $1.95 \times 10^{7}$ to $2.52 \times 10^{2}$---surpassing both components.

The green entries in \cref{tab:mixed_alpha_sensitivity_exp_colored} reveal that nearly all LSTM-involved combinations (either LSTM first or second) beat both standalone baselines consistently across all memory functions.
In contrast, non-LSTM combinations show memory-function-dependent results: some beat both baselines (e.g., Transformer$\leftrightarrow$S4 in Airy, S4$\rightarrow$TCN in Impulse), while others beat only one baseline or neither.
This difference arises because LSTM is consistently weak across all memory functions, creating universal opportunities for improvement, whereas other architectures exhibit memory-function specific strengths and weaknesses.
Large performance gaps enable synergy: the weaker architecture benefits from processing intermediate features rather than the raw target directly, while the stronger architecture can effectively leverage even imperfect features from the weaker component.

Ordering effects reveal asymmetric synergies.
For exponential and polynomial functions, placing S4 or Transformer as the second stage consistently yields the best results, suggesting these architectures are more effective at refining features extracted by earlier layers.
For the impulse function, TCN must be the second stage to leverage its convolutional receptive field for sparse long-range dependencies; placing TCN first provides no benefit.
This asymmetry indicates that architectural positioning matters: feature extraction and feature refinement require different inductive biases. 
 
These findings suggest mixed architectures as a practical strategy for handling diverse temporal structures when dependency patterns are unknown, allowing complementary strengths to compensate for individual weaknesses.
From a theoretical perspective, we conjecture that sequential composition can achieve improved approximation guarantees compared to individual architectures: if the first architecture transforms the input into an intermediate representation that emphasizes temporal features suited to the second architecture's inductive bias, the composed model may attain better approximation rates than either component alone.
The results motivate future work on principled methods for selecting architecture orderings based on memory function characteristics.

\begin{table}[hbtp]
    \centering
    \caption{Sensitivity to Temporal Strength for Mixed Models - EXP Memory Function.
    Relative Loss Increment from Minimum to Maximum $\alpha$ (Mean $\pm$ SEM).
    Rows: first architecture, columns: second architecture.
    Diagonal (gray): single-architecture baselines; green: improvement over both baselines; light green: improvement over one baseline; red: degradation below both baselines.}
    \label{tab:mixed_alpha_sensitivity_exp_colored}
    \small 
    \begin{tabular}{lcccc}
    \hline
      & LSTM & S4 & TCN & Transformer \\
    \hline
    LSTM & \cellcolor{gray!20}$1.32 \pm 1.27 \times 10^{1}$ & \cellcolor{green!90!white}$1.74 \pm 0.70 \times 10^{-1}$ & \cellcolor{green!90!white}$1.14 \pm 0.25 \times 10^{0}$ & \cellcolor{green!90!white}$1.37 \pm 0.21 \times 10^{-1}$ \\
    S4 & \cellcolor{green!90!white}$7.21 \pm 3.26 \times 10^{-2}$ & \cellcolor{gray!20}$2.17 \pm 0.77 \times 10^{-1}$ & \cellcolor{green!25!white}$4.50 \pm 0.94 \times 10^{-1}$ & \cellcolor{green!25!white}$3.21 \pm 0.70 \times 10^{-1}$ \\
    TCN & \cellcolor{green!90!white}$3.21 \pm 2.06 \times 10^{0}$ & \cellcolor{green!25!white}$8.91 \pm 4.12 \times 10^{-1}$ & \cellcolor{gray!20}$1.04 \pm 0.20 \times 10^{1}$ & \cellcolor{green!25!white}$2.91 \pm 0.33 \times 10^{1}$ \\
    Transformer & \cellcolor{green!90!white}$7.67 \pm 0.96 \times 10^{-1}$ & \cellcolor{green!25!white}$1.72 \pm 0.20 \times 10^{0}$ & \cellcolor{red!30!white}$1.36 \pm 1.02 \times 10^{2}$ & \cellcolor{gray!20}$1.01 \pm 0.13 \times 10^{2}$ \\
    \hline
    \end{tabular}
\end{table}

\revision{
\subsection{Mixture of Targets}
\label{subsec:mixed_target}

The preceding sections evaluate each memory function in isolation,
which enables precise characterization of architectural strengths and
weaknesses with respect to specific temporal structures. However,
practical sequence modeling tasks often involve targets with hybrid
temporal dependencies, where multiple distinct memory structures are
present simultaneously. To investigate whether the architectural
trade-offs identified in \cref{subsec:recurrent}--\cref{subsec:transformer}
persist under such conditions, we extend our evaluation to mixed
memory function targets.

We define a mixed target by linearly combining two parametric memory
functions:
\begin{equation}
    \rho_{\text{mix}}(s, \alpha, \beta)
    = (1 - \beta)\,\rho_1(s, \alpha) + \beta\,\rho_2(s, \alpha),
    \quad \beta \in [0, 1],
\end{equation}
where $\beta$ controls the relative contribution of each component.
Since both $\rho_1$ and $\rho_2$ are valid memory functions satisfying
the conditions of \cref{thm: linear target representation}, their linear combination
$\rho_{\text{mix}}$ is also a valid memory function, and the resulting
synthetic target retains the same controlled structure as the
single-kernel cases.

We focus on the combination of exponential and impulse memory functions
with $\alpha = 1$,
\begin{equation}
    \rho_{\text{mix}}(s, \beta)
    = (1 - \beta)\,\rho_{\exp}(s, 1) + \beta\,\rho_{\delta}(s, 1).
\end{equation}
This combination is chosen because $\rho_{\exp}$ and $\rho_{\delta}$
represent the most architecturally contrasting temporal structures in
our benchmark, with recurrent architectures performing best on the
former and convolutional architectures performing best on the latter
(\cref{tab:alpha_sensitivity}). Their mixture therefore creates a
target where no single architecture is naturally suited to both
components simultaneously, making it an effective probe of
architectural robustness under hybrid temporal dependencies. We vary
$\beta$ from $0$ (pure exponential) to $1$ (pure impulse) and measure
approximation loss for each architecture.
\begin{figure}[htbp] 
    \centering
    \includegraphics[width=0.7\linewidth]{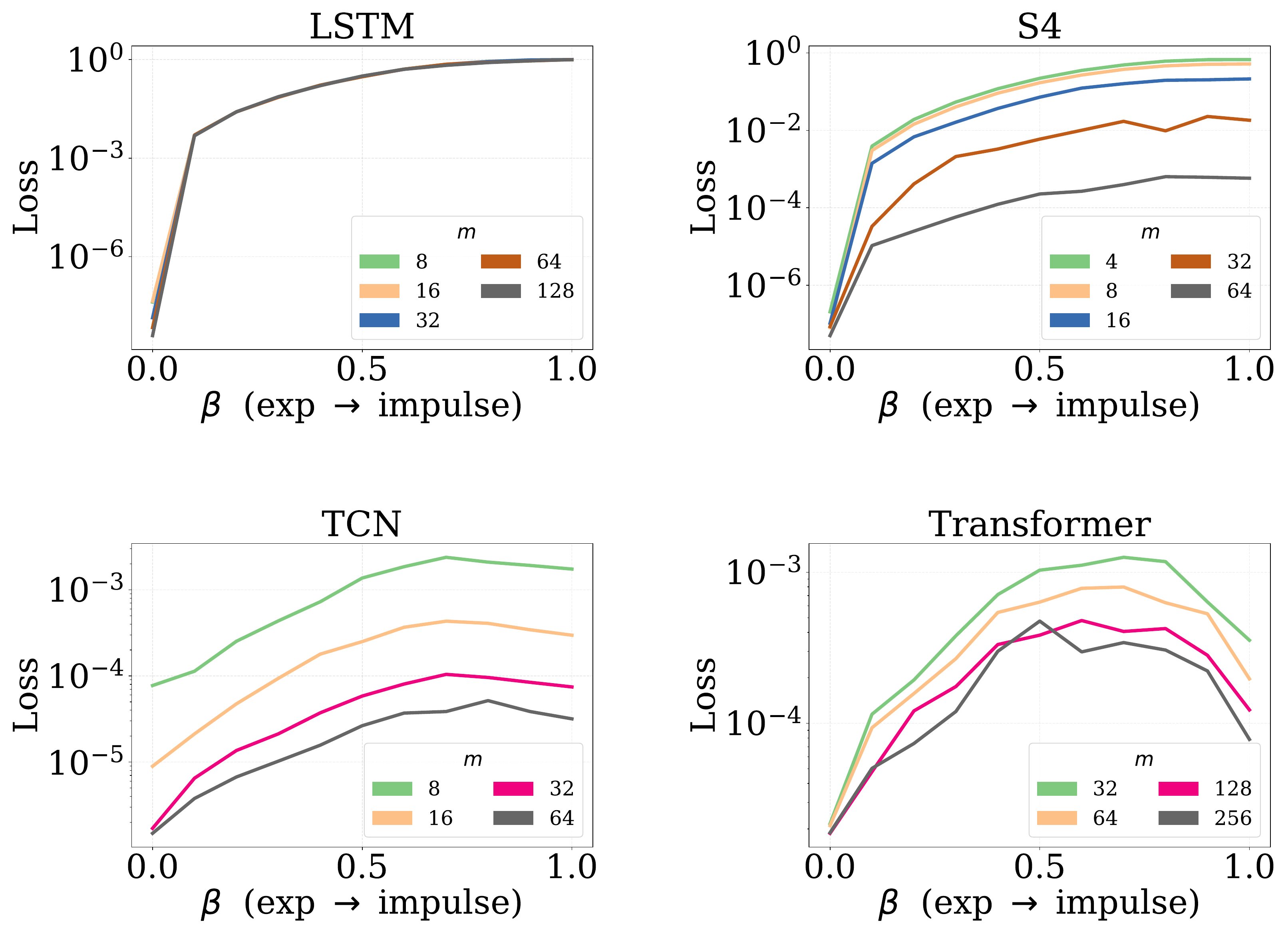}
    \caption{Loss versus $\beta$ for all four architectures on the mixed
    memory function $\rho_{\text{mix}}(s, \beta) = (1-\beta)\,\rho_{\exp}(s, 1)
    + \beta\,\rho_{\delta}(s, 1)$, with varying model sizes $m$.
    LSTM collapses immediately at small $\beta$ with no model size benefit.
    S4D degrades sharply but with some separation across model sizes.
    TCN degrades gradually with clear model size separation throughout.
    The Transformer exhibits non-monotonic behavior, peaking near $\beta \approx 0.5$
    before recovering toward $\beta = 1$, with the largest benefit from increased model size.}
    \label{fig:mixed_target}
\end{figure}


As shown in \cref{fig:mixed_target}, the rate and pattern of degradation
across architectures differ substantially as $\beta$ increases,
revealing trade-offs not visible in the single-kernel evaluations.

LSTM and S4D both achieve low loss at $\beta = 0$, consistent with
their strong performance on the exponential memory function
(\cref{tab:alpha_sensitivity}). However, both degrade sharply even for
small $\beta$, with loss saturating near $10^{0}$ and $10^{-2}$. This confirms that
recurrent architectures cannot tolerate even a modest impulse
component: the long-range sparse dependency fundamentally disrupts
their approximation mechanism, as established in
\cref{subsec:recurrent} and supported by
\cite{bengio1994.LearningLongtermDependencies,
li2022.ApproximationOptimizationTheory,
wang2023.InverseApproximationTheory}.

TCN degrades more gradually than LSTM and S4D.
Importantly, TCN exhibits clear and consistent model size separation
throughout the full $\beta$ range, where larger $m$ yields substantially lower
loss at every mixture ratio,
indicating that increased capacity provides a meaningful and sustained benefit
across all mixture ratios. 

The Transformer exhibits qualitatively different behavior from all other
architectures: its loss is non-monotonic in $\beta$, peaking near
$\beta \approx 0.5$ before recovering as $\beta \to 1$. This indicates
that the Transformer handles the pure impulse target more effectively
than intermediate mixtures, where both components must be approximated
simultaneously. The recovery at large $\beta$ suggests that as the
exponential component becomes negligible, the Transformer can
concentrate its attention capacity on the sparse long-range dependency,
recovering close to its single-kernel impulse performance. 

 
These results suggest that architectural specialization, while
beneficial for targets with well-defined single temporal structures,
may become a disadvantage when dependencies are mixed. The
Transformer's moderate but consistent performance across individual
memory functions translates into greater robustness under hybrid
dependencies. This offers a complementary strategy to the sequential
composition approach studied in \cref{subsec:mixed}, achieving
robustness through breadth of approximation capacity rather than
through explicit architectural combination.
}
\section{Conclusion}\label{sec:conclusion}

In this work, we introduced a systematic benchmarking framework for evaluating sequence modeling architectures using parametric memory functions $\rho(s, \alpha)$ with controllable temporal dependencies.
By varying the parameter $\alpha$, our approach generates a continuum of synthetic targets that isolate specific temporal properties, enabling fine-grained analysis of architectural capabilities.
Experiments on LSTM, S4D, TCN, and Transformer architectures validated existing theoretical predictions while revealing new phenomena.
We confirmed that recurrent and state-space models handle exponential decay effectively but struggle with polynomial decay and sparse long-range dependencies, with depth effects depending critically on memory function structure.
For TCNs, we introduced a novel tail energy complexity measure that characterizes approximation difficulty more precisely than binary sparsity criteria.
For Transformers, we validated effective rank theory and identified critical trade-offs between attention head count and per-head dimensionality that depend on temporal structure.
Sequential composition experiments demonstrated that large performance gaps between architectures enable synergies, with weaker architectures benefiting from processing intermediate features rather than raw targets.

While our analysis focused primarily on approximation properties with scalar targets, the framework naturally accommodates extensions to optimization and generalization analysis, multivariate sequences, and additional memory function classes.
Future work could explore bridging the gap between synthetic evaluation and real-world performance by developing methods to infer memory function properties from data or predict architecture suitability based on dataset characteristics.
Our results demonstrate that no single architecture excels across all temporal structures, highlighting the importance of understanding temporal dependencies when selecting architectures for practical applications.
As sequence modeling continues to evolve, systematic evaluation frameworks like the one proposed here will play an increasingly important role in characterizing capabilities, identifying limitations, and guiding architectural design.

\appendix
\crefalias{section}{appendix}

\section{Proofs and technical details} \label{appen: proof}

\subsection{Proof of Theorem~\ref{thm: linear target representation}}
\begin{proof} 
Firstly, recall that the \normalfont{Riesz Representation Theorem} for Hilbert space states that for any continuous linear functional $H$ defined on the Hilbert space $\tilde \X := \{ \bm x \in \X: \displaystyle\sum\abs{x}^2<\infty \}$,
there exists a unique $\rho \in \tilde\X$
such that
	 \begin{equation}
	 	H(\bm x) = \sum_{s = 0}^{\infty} \rho(s) x(s),
	 \end{equation}
and
\begin{equation}
    \norm{H} = \norm{\rho}_{\tilde\X}.
\end{equation}
This is a classic result that can be found for example in \cite{kreyszig1989.IntroductoryFunctionalAnalysis}, Theorem 3.8-1.

By this Riesz Representation Theorem, we have that for any $t \in \mathbb Z$ and $H_t \in \bm H$, there exists a unique $\bm \rho_t \in \tilde\X$ such that
    \begin{equation}
        H_t(\bm x) = \sum_{s=0}^{\infty}\rho_t(s) x(s).
    \end{equation}
    With the fact that $\bm H$ is causal, we have
    \begin{equation}
        H_t(\bm x) = \sum_{s=0}^{t}\rho_t(s) x(s).
    \end{equation}
    By the time homogeneity $H_t(\bm x) = H_{t+\tau}(\bm x\depen{\tau})$, we have:
    \begin{align}
        H_t(\bm x) = \sum_{s=0}^t\rho_t(s)x(s) &= \sum_{s=\tau}^{t+\tau}\rho_{t+\tau}(s)x(s-\tau), \\
        & = \sum_{s=0}^{t}\rho_{t+\tau}(s+\tau)x(s) ,
    \intertext{   Let $\tau=t-2s$ we get,}
        & = \sum_{s=0}^{t}\rho_{2(t-s)}(t-s)x(s).
   \end{align}
   Let $\rho\depen{\bm H}(s) = \rho_{2s}(s)$, we get
   \begin{equation}
    	H_t(\bm x) = \sum_{s=0}^{t} \rho\depen{\bm H}(s) x(t-s) .
    \end{equation}

\end{proof}

\subsection{TCN Approximation Theorem}

We present a modified approximation bound for Temporal Convolutional Networks (TCNs) that incorporates the tail energy complexity measure introduced in \cref{sec:results}.
The theorem characterizes how well a TCN with $K$ layers and filter size $l$ can approximate a target function with memory structure $\rho$.

\paragraph{TCN Architecture Definition}
A convolution based temporal sequence model with $K$ layers and $M_k$ channels at layer $k$ is given by
\begin{equation}\label{CNNdynamics}
        \begin{aligned}
        \bm h_{0,i} &= \bm x_i,\\
        \bm h_{k+1,i} &= \sigma \left(\sum_{j=1}^{M_k} {\bm w}_{kji} \ast_{d_k} \bm h_{k,j} \right),\\
        \bm {\hat y} &= \bm h_K.
        \end{aligned}
\end{equation}
Here, $\bm x_{i}$ is the $i^{th}$ dimension of $\bm x$, and
${\bm w}_{kji}$ is the filter from channel $j$ at layer $k$ to channel $i$ at layer $k+1$.
All the filters have a size $l\geq2$, i.e. $r({\bm w}_{kji}) = l \ge 2$.
Let $M = \sum_{k=1}^{K-1} M_kM_{k-1} - K$ denote the effective number of filters used in the model.
Here we minus the number of layers $K$ because each layer has at least one filter, which also implies $M \geq K$.
In the following discussion, we assume that the dilation rate satisfies $d_k = l^k$.
That is, the dilation rate increases exponentially for each layer to achieve
an exponentially large receptive field.
This is the standard practice for dilated convolutional structures \cite{oord2016.WaveNetGenerativeModel,yu2016.MultiScaleContextAggregation}.

\textbf{Notation:}
Let $K$ denote the number of TCN layers, and $M_k$ denote the number of channels in layer $k$.
The TCN architecture has a receptive field of size $l^K$, where $l$ is the filter size.
We denote by $C\depen{l,g}$ the space of target functions with filter size $l$ and growth condition $g$,
and by $\mathcal H_{\text{CNN}}\depen{l,K,\{M_k\}}$ the class of functions representable by TCNs with the given architecture.
For a target function $\bm H$ with memory function $\rho$,
we denote by $\hat{\bm H}$ its TCN approximation with memory function $\hat{\rho}$,
and by $\rho_{[l^K,\infty]}$ the tail of the memory function beyond the receptive field.
The permutation $\pi$ reorders the memory function values in non-increasing order, as defined in \cref{eq: complexity measure}.

\begin{theorem}\label{thm: tcn approximation}
Fix the filter size $l\in\mathbb N_+$.
Let
$M := \sum_{k=2}^K M_kM_{k-1} - K$ be the effective number of filters.
Then for any set of parameters
$(K,\{M_k\})$,
and any $\bm H  \in C\depen{l,g}$, we have
\begin{equation}\label{errorbound}
	\begin{aligned}
		\inf_{\hat{\bm H} \in \mathcal H_{\text{CNN}}\depen{l,K,\{M_k\}}}\norm{\bm H - \hat{\bm H}}
		 \leq
		 d\,  C(\rho, M+1)
		   + { d\norm{\rho_{[l^K,\infty]}}_2},
	\end{aligned}
\end{equation}
where
\begin{align}
    C(\rho, s) := \sum_{t=s}^{l^K-1} \abs{\rho(\pi(t))}^2.
\end{align}
\end{theorem}

\begin{proof}
This is a direct consequence of Equation (18) from \cite{jiang2021.ApproximationTheoryConvolutional}:
\begin{align}
    \abs{H_t(\bm x)  - \hat H_t(\bm x)} & \leq
    \sum_{s=0}^\infty \Big\rvert \rho(s) - \hat{\rho}(s) \Big\rvert \\
     &= \sum_{s=0}^{l^K-1}\Big\rvert \rho(s) - \hat{\rho}(s)\Big\rvert
    + \sum_{s=l^K}^{\infty} \Big\rvert \rho(s) \Big\rvert. \nonumber
\end{align}
For the first term we have
\begin{align}
     \sum_{s=0}^{l^K-1}\Big\rvert \rho(s) - \hat{\rho}(s)\Big\rvert &= \sum_{s=0}^{l^K-1}\Big\rvert \rho(\pi(s)) - \hat{\rho}(\pi(s))\Big\rvert \\
    & = C(\rho, M+1).
\end{align}
We use the $M$ channels to cover the largest $M$ values of $\rho$, and the remainder error is characterized by $C(\rho, M+1)$.
\end{proof}

\section{Sequence Length Ablation}\label{appendix:length_ablation}

This appendix presents the full length ablation results, evaluating LSTM, S4D, and TCN at $T \in \{512, 2048\}$ and Transformer at $T \in \{64, 256\}$.
The relative difficulty ordering across memory functions and the sensitivity patterns reported in \cref{sec:results} remain consistent across all tested lengths, confirming that the observed trends are not an artifact of the chosen sequence lengths.

\begin{figure}[!ht]
    \centering
    \includegraphics[width=0.85\linewidth]{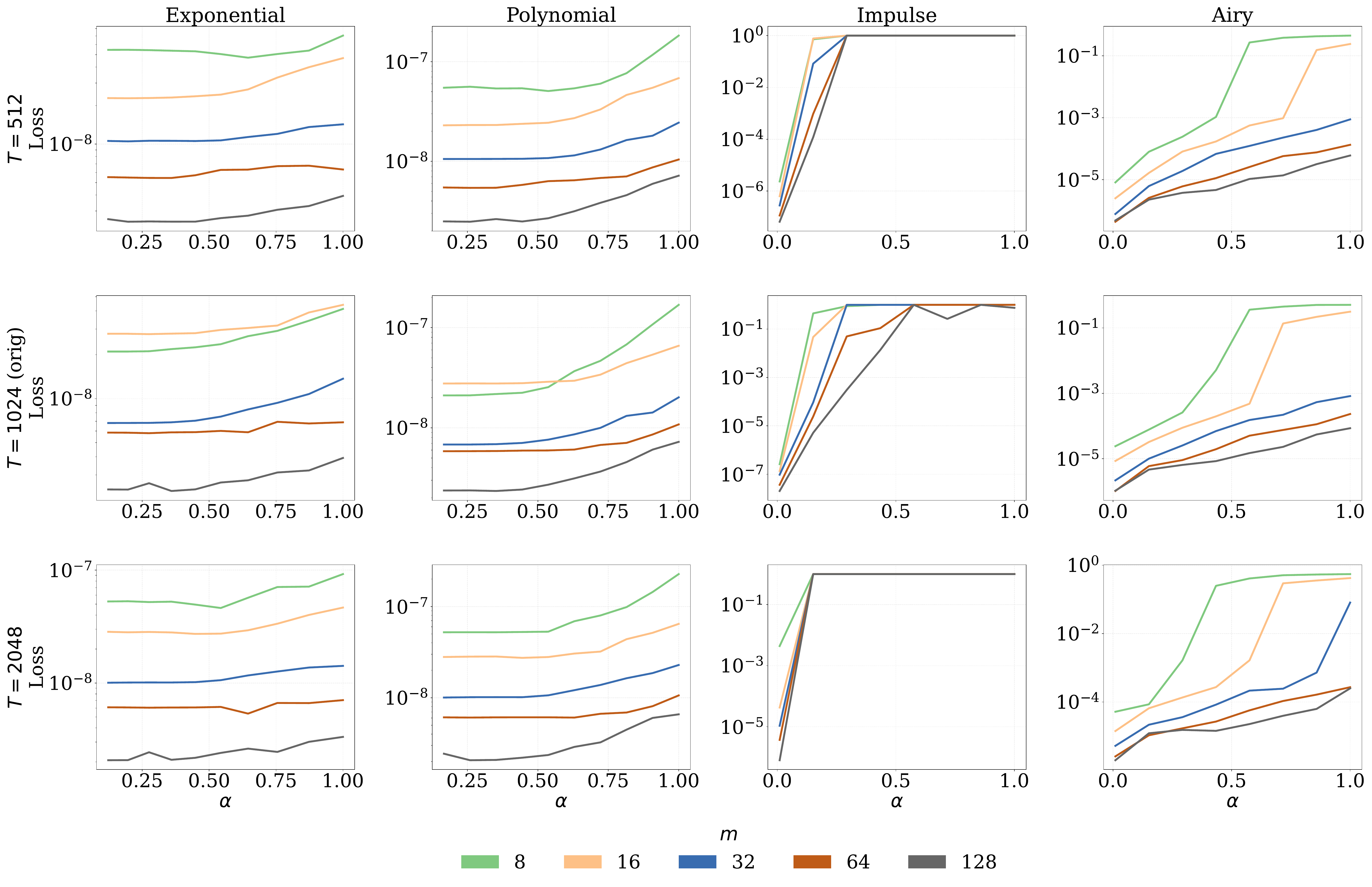}
    \caption{Length ablation for LSTM at $T \in \{512, 2048\}$ across all four memory functions.}
    \label{fig:length_ablation_lstm}
\end{figure}

\begin{figure}[!ht]
    \centering
    \includegraphics[width=0.85\linewidth]{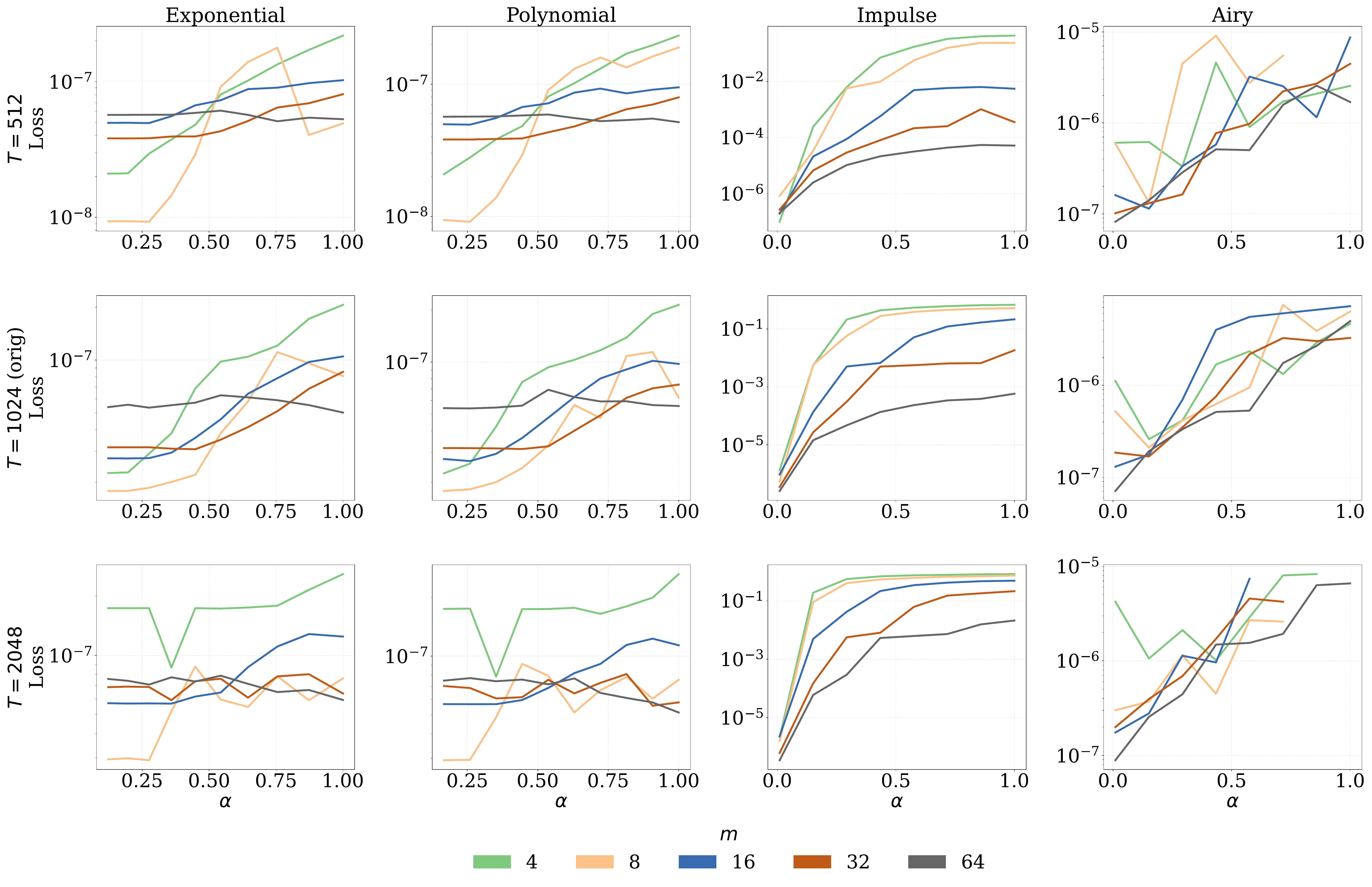}
    \caption{Length ablation for S4D at $T \in \{512, 2048\}$ across all four memory functions.}
    \label{fig:length_ablation_s4}
\end{figure}

\begin{figure}[!ht]
    \centering
    \includegraphics[width=0.85\linewidth]{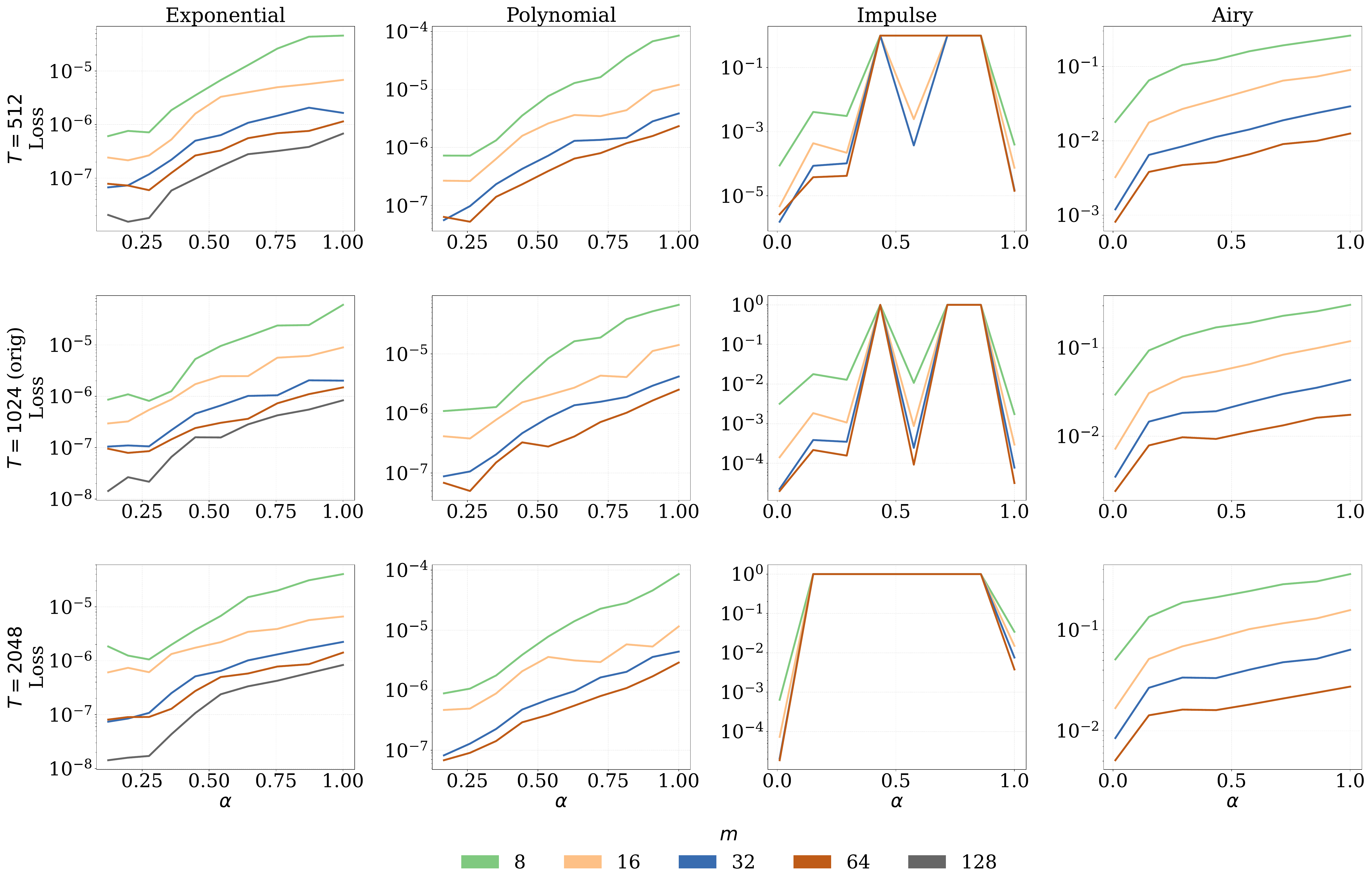}
    \caption{Length ablation for TCN at $T \in \{512, 2048\}$ across all four memory functions.}
    \label{fig:length_ablation_tcn}
\end{figure}

\begin{figure}[!ht]
    \centering
    \includegraphics[width=0.85\linewidth]{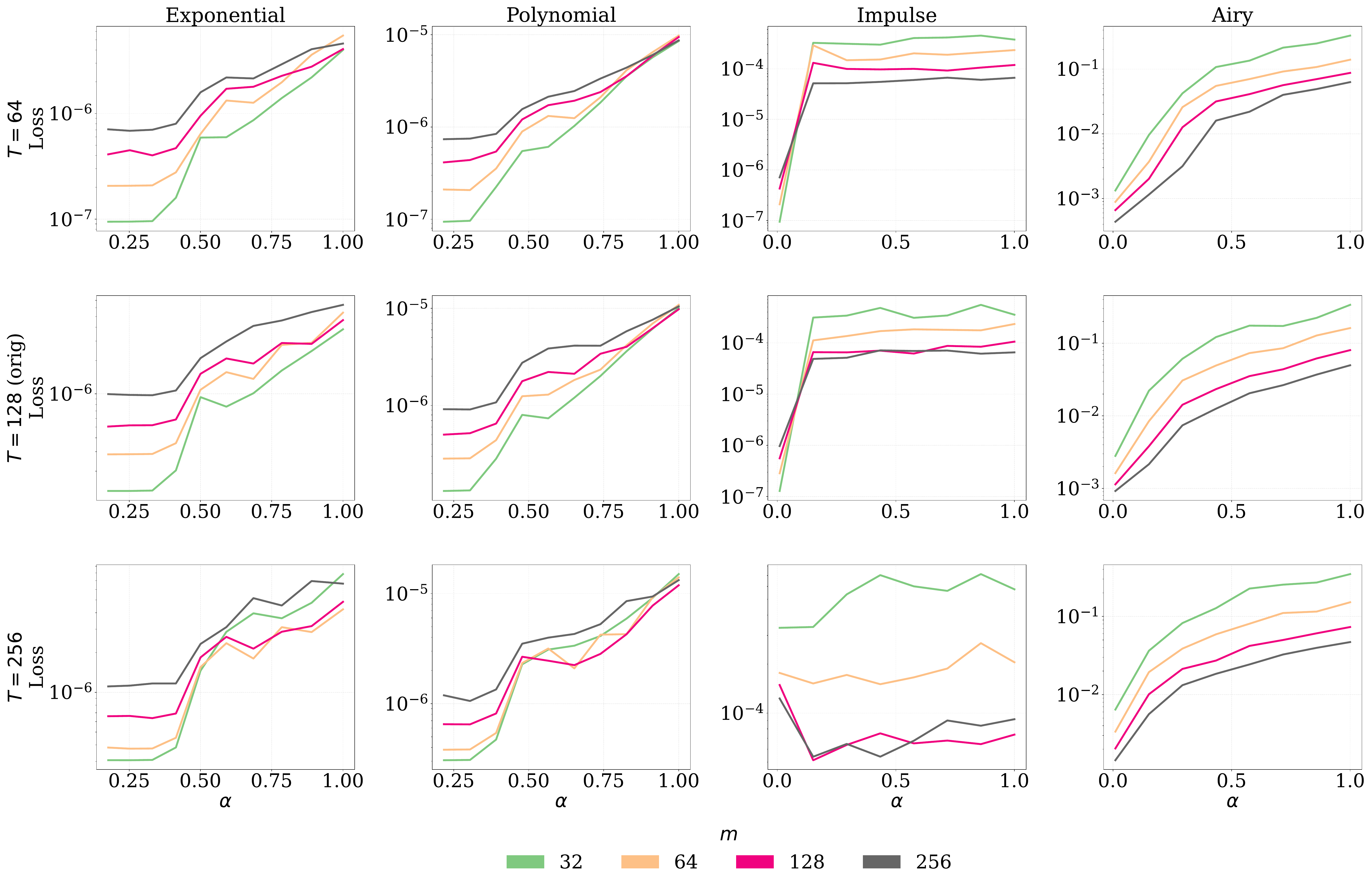}
    \caption{Length ablation for Transformer at $T \in \{64, 256\}$ across all four memory functions.}
    \label{fig:length_ablation_transformer}
\end{figure}

\section{Train-Validation Consistency}\label{appendix:val_curves}

This appendix presents validation loss curves for all four architectures across all four memory functions and model sizes.
In each plot, validation loss follows the same trend as training loss, confirming that no significant overfitting occurs across the parameter settings used in this study.
The consistency between training and validation losses supports the use of final training loss as a reliable proxy for approximation error.

\begin{figure}[!ht]
    \centering
    \includegraphics[width=0.7\linewidth]{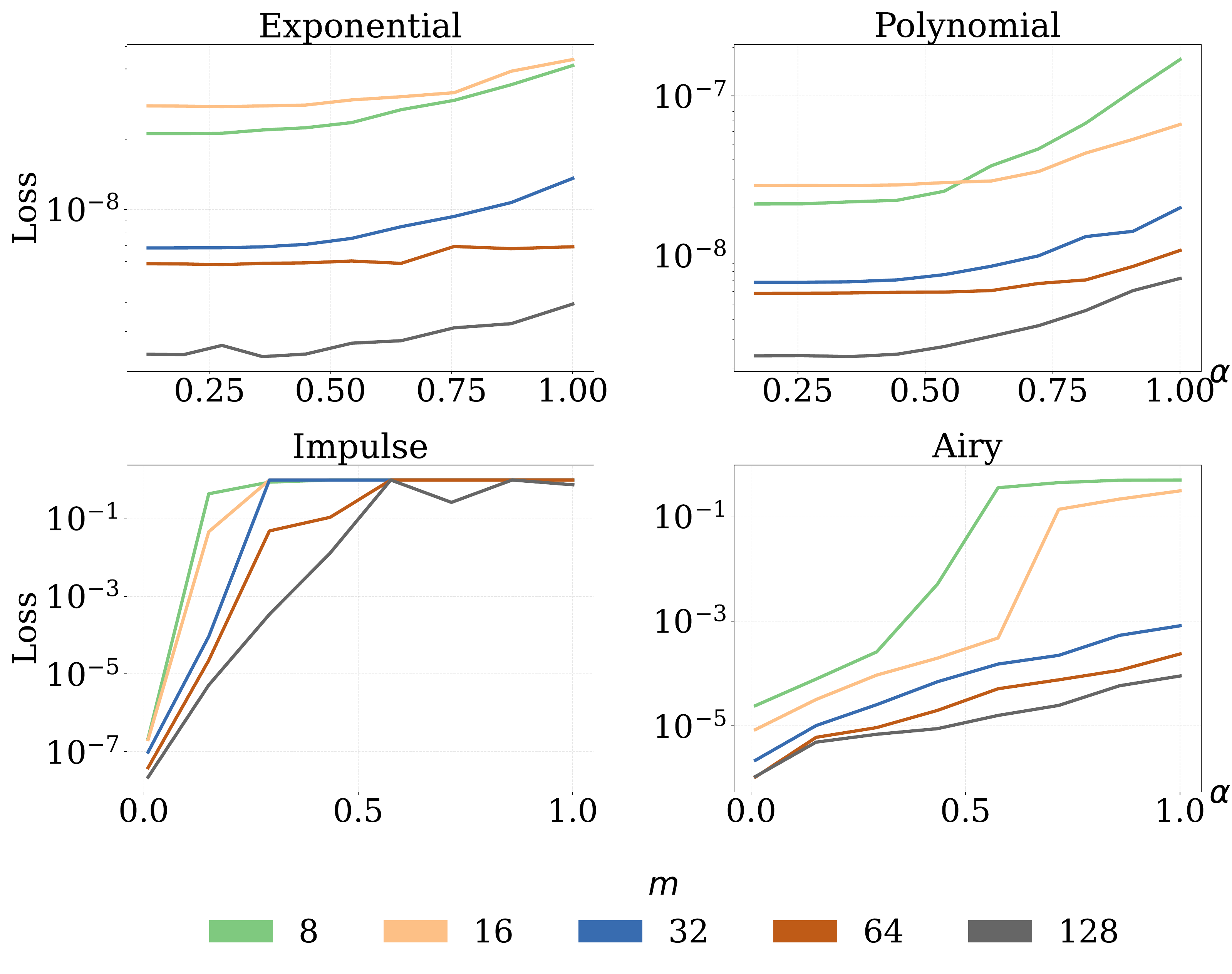}
    \caption{Validation loss for LSTM across all four memory functions and model sizes $m \in \{8, 16, 32, 64, 128\}$.}
    \label{fig:val_lstm}
\end{figure}

\begin{figure}[!ht]
    \centering
    \includegraphics[width=0.7\linewidth]{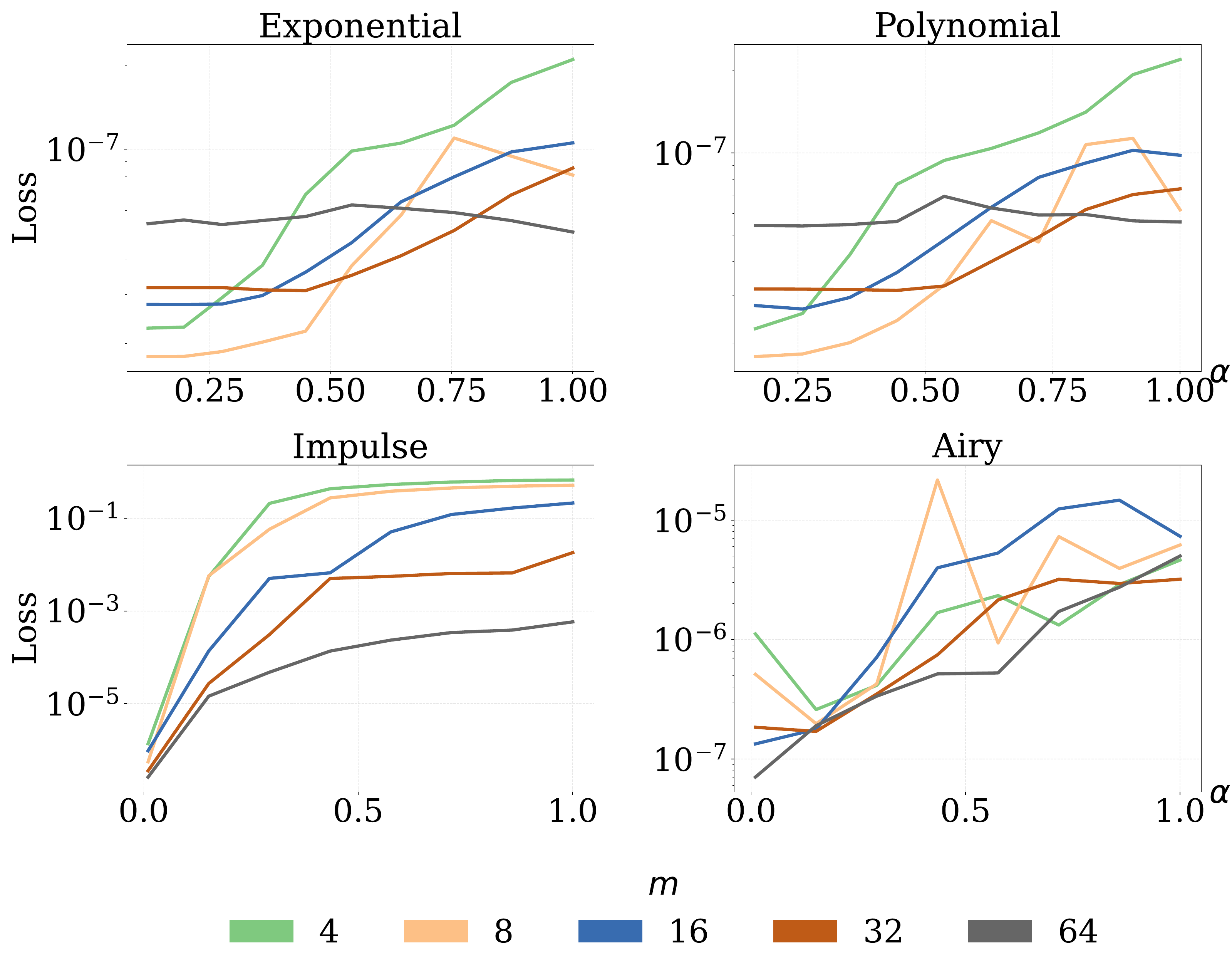}
    \caption{Validation loss for S4D across all four memory functions and model sizes $m \in \{4, 8, 16, 32, 64\}$.}
    \label{fig:val_s4}
\end{figure}

\begin{figure}[!ht]
    \centering
    \includegraphics[width=0.7\linewidth]{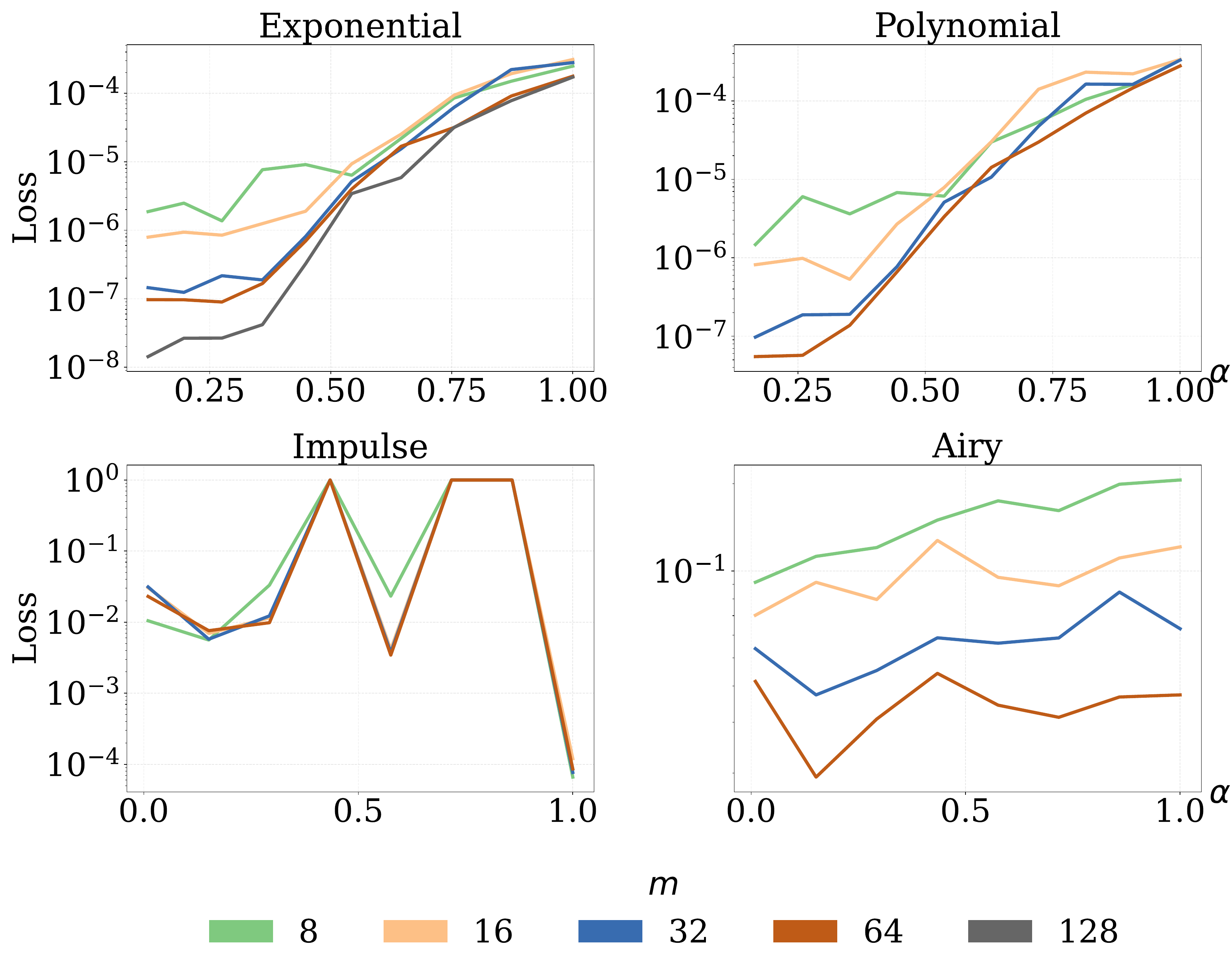}
    \caption{Validation loss for TCN across all four memory functions and model sizes $m \in \{8, 16, 32, 64, 128\}$.}
    \label{fig:val_tcn}
\end{figure}

\begin{figure}[!ht]
    \centering
    \includegraphics[width=0.7\linewidth]{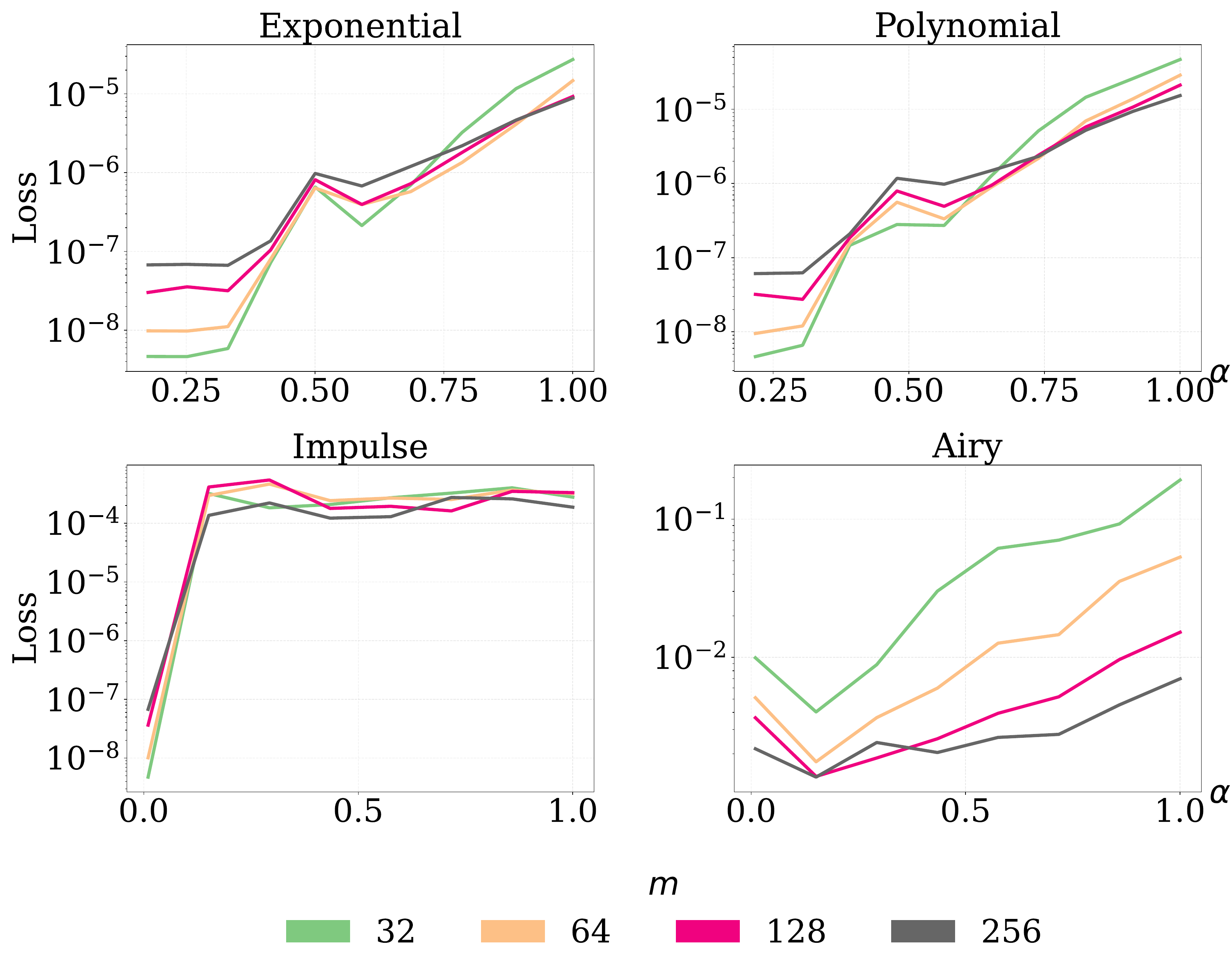}
    \caption{Validation loss for Transformer across all four memory functions and model sizes $m \in \{32, 64, 128, 256\}$.}
    \label{fig:val_transformer}
\end{figure}

\section{Additional Mixed Model Results}\label{appendix:mixed_tables}

This appendix presents detailed alpha sensitivity results for mixed model architectures under polynomial, impulse, and Airy memory functions.
The exponential memory function results are presented in the main text (\cref{tab:mixed_alpha_sensitivity_exp_colored}).
All tables show the relative loss increment from minimum to maximum $\alpha$ (Mean $\pm$ SEM) for pairwise combinations of LSTM, S4, TCN, and Transformer architectures.
Rows indicate the first architecture in the composition, columns indicate the second.
Diagonal entries (gray) represent single-architecture baselines.
Green shading indicates improvement over both baselines, light green indicates improvement over one baseline, while red indicates degradation below both baselines.

\begin{table}[hbtp]
    \centering
    \caption{Sensitivity to Temporal Strength for Mixed Models - POLY Memory Function: Relative Loss Increment from Minimum to Maximum $\alpha$ (Mean $\pm$ SEM). Rows: first architecture. Columns: second architecture. Diagonal (gray): single-architecture baselines. Green: improvement over both baselines. Light green: improvement over one baseline. Red: degradation below both baselines.}
    \label{tab:mixed_alpha_sensitivity_poly_colored}
    \small
    \begin{tabular}{lcccc}
    \hline
     & LSTM & S4 & TCN & Transformer \\
    \hline
    LSTM & \cellcolor{gray!20}$2.48 \pm 2.09 \times 10^{0}$ & \cellcolor{green!90!white}$1.72 \pm 0.53 \times 10^{-1}$ & \cellcolor{green!90!white}$2.36 \pm 0.64 \times 10^{0}$ & \cellcolor{green!90!white}$8.86 \pm 4.08 \times 10^{-1}$ \\
    S4 & \cellcolor{green!90!white}$7.41 \pm 2.99 \times 10^{-2}$ & \cellcolor{gray!20}$1.96 \pm 0.35 \times 10^{-1}$ & \cellcolor{green!25!white}$3.74 \pm 0.80 \times 10^{-1}$ & \cellcolor{green!25!white}$2.84 \pm 0.73 \times 10^{-1}$ \\
    TCN & \cellcolor{green!90!white}$1.56 \pm 0.50 \times 10^{0}$ & \cellcolor{green!25!white}$8.93 \pm 3.34 \times 10^{-1}$ & \cellcolor{gray!20}$2.77 \pm 1.72 \times 10^{1}$ & \cellcolor{green!25!white}$8.18 \pm 3.27 \times 10^{1}$ \\
    Transformer & \cellcolor{green!90!white}$1.57 \pm 0.21 \times 10^{0}$ & \cellcolor{green!25!white}$2.25 \pm 0.30 \times 10^{0}$ & \cellcolor{green!25!white}$1.08 \pm 0.63 \times 10^{2}$ & \cellcolor{gray!20}$1.68 \pm 0.17 \times 10^{2}$ \\
    \hline
    \end{tabular}
\end{table}

\begin{table}[hbtp]
    \centering
    \caption{Sensitivity to Temporal Strength for Mixed Models - IMPULSE Memory Function: Relative Loss Increment from Minimum to Maximum $\alpha$ (Mean $\pm$ SEM). Rows: first architecture. Columns: second architecture. Diagonal (gray): single-architecture baselines. Green: improvement over both baselines. Light green: improvement over one baseline. Red: degradation below both baselines.}
    \label{tab:mixed_alpha_sensitivity_impulse_colored}
    \small
    \begin{tabular}{lcccc}
    \hline
     & LSTM & S4 & TCN & Transformer \\
    \hline
    LSTM & \cellcolor{gray!20}$1.95 \pm 0.46 \times 10^{7}$ & \cellcolor{green!90!white}$3.72 \pm 2.11 \times 10^{3}$ & \cellcolor{green!90!white}$2.52 \pm 0.67 \times 10^{2}$ & \cellcolor{green!90!white}$1.44 \pm 0.44 \times 10^{4}$ \\
    S4 & \cellcolor{green!90!white}$9.22 \pm 6.22 \times 10^{3}$ & \cellcolor{gray!20}$9.68 \pm 3.28 \times 10^{1}$ & \cellcolor{green!90!white}$1.10 \pm 0.30 \times 10^{1}$ & \cellcolor{red!30!white}$3.98 \pm 2.04 \times 10^{3}$ \\
    TCN & \cellcolor{green!90!white}$7.59 \pm 1.16 \times 10^{1}$ & \cellcolor{green!25!white}$5.27 \pm 1.92 \times 10^{1}$ & \cellcolor{gray!20}$-1.01 \pm 2.06 \times 10^{-1}$ & \cellcolor{green!90!white}$-3.17 \pm 0.82 \times 10^{-1}$ \\
    Transformer & \cellcolor{green!90!white}$2.88 \pm 0.73 \times 10^{3}$ & \cellcolor{red!30!white}$5.64 \pm 5.09 \times 10^{2}$ & \cellcolor{green!25!white}$6.98 \pm 2.17 \times 10^{-1}$ & \cellcolor{gray!20}$4.27 \pm 2.66 \times 10^{1}$ \\
    \hline
    \end{tabular}
\end{table}

\begin{table}[hbtp]
    \centering
    \caption{Sensitivity to Temporal Strength for Mixed Models - AIRY Memory Function: Relative Loss Increment from Minimum to Maximum $\alpha$ (Mean $\pm$ SEM). Rows: first architecture. Columns: second architecture. Diagonal (gray): single-architecture baselines. Green: improvement over both baselines. Light green: improvement over one baseline. Red: degradation below both baselines.}
    \label{tab:mixed_alpha_sensitivity_airy_colored}
    \small
    \begin{tabular}{lcccc}
    \hline
   & LSTM & S4 & TCN & Transformer \\
    \hline
    LSTM & \cellcolor{gray!20}$1.49 \pm 0.50 \times 10^{4}$ & \cellcolor{green!90!white}$1.34 \pm 0.23 \times 10^{1}$ & \cellcolor{green!90!white}$1.05 \pm 0.21 \times 10^{3}$ & \cellcolor{green!90!white}$4.79 \pm 1.22 \times 10^{3}$ \\
    S4 & \cellcolor{green!90!white}$1.31 \pm 0.21 \times 10^{1}$ & \cellcolor{gray!20}$2.02 \pm 0.38 \times 10^{1}$ & \cellcolor{red!30!white}$2.81 \pm 0.86 \times 10^{1}$ & \cellcolor{green!90!white}$7.70 \pm 1.65 \times 10^{0}$ \\
    TCN & \cellcolor{green!90!white}$1.80 \pm 0.19 \times 10^{3}$ & \cellcolor{red!30!white}$6.87 \pm 1.78 \times 10^{1}$ & \cellcolor{gray!20}$1.32 \pm 0.17 \times 10^{1}$ & \cellcolor{green!25!white}$1.34 \pm 0.10 \times 10^{1}$ \\
    Transformer & \cellcolor{green!90!white}$2.22 \pm 0.27 \times 10^{3}$ & \cellcolor{green!90!white}$1.91 \pm 0.38 \times 10^{0}$ & \cellcolor{green!90!white}$1.1 \pm 0.26 \times 10^{1}$ & \cellcolor{gray!20}$1.45 \pm 0.25 \times 10^{1}$ \\
    \hline
    \end{tabular}
\end{table}

\newpage
\thispagestyle{plain} 
\mbox{}
\clearpage

\bibliographystyle{unsrt}
\bibliography{ref.bib}

\clearpage

\end{document}